\documentclass{article}

\usepackage{microtype}
\usepackage{graphicx}
\usepackage{subfigure}
\usepackage{booktabs} 

\usepackage{amsmath,amsfonts,bm}

\def\1{\bm{1}}

\DeclareMathAlphabet{\mathsfit}{\encodingdefault}{\sfdefault}{m}{sl}
\SetMathAlphabet{\mathsfit}{bold}{\encodingdefault}{\sfdefault}{bx}{n}

\DeclareMathOperator*{\argmin}{arg\,min}

\usepackage{hyperref}

\usepackage[accepted]{icml2024}

\usepackage{amsmath}
\usepackage{amssymb}
\usepackage{mathtools}
\usepackage{amsthm}

\usepackage[capitalize,noabbrev]{cleveref}

\theoremstyle{plain}

\theoremstyle{definition}

\theoremstyle{remark}

\newcommand{\beq}{ \begin{equation} }
\newcommand{\eeq}{ \end{equation} }
\usepackage[textsize=tiny]{todonotes}

\usepackage{multirow}
\usepackage{threeparttable}
\usepackage{pifont}
\usepackage{adjustbox}

\newcommand{\DAG}{\textsuperscript{\textdagger}}

\newcommand{\xmark}{\ding{55}}

 \icmltitlerunning{SelMatch: Effectively Scaling Up Dataset Distillation via Selection-Based Initialization and Partial Updates}

\begin{document}


\twocolumn[
\icmltitle{SelMatch: Effectively Scaling Up Dataset Distillation via Selection-Based Initialization and Partial Updates by Trajectory Matching}

\begin{icmlauthorlist}
\icmlauthor{Yongmin Lee}{kaist}
\icmlauthor{Hye Won Chung}{kaist}
\end{icmlauthorlist}

\icmlaffiliation{kaist}{School of Electrical Engineering, Korea Advanced Institute of Science and Technology (KAIST), Daejeon, South Korea}

\icmlcorrespondingauthor{Yongmin Lee}{lym7505@kaist.ac.kr}
\icmlcorrespondingauthor{Hye Won Chung}{hwchung@kaist.ac.kr}

\icmlkeywords{Dataset Distillation, Subset Selection, Data-efficient Learning}

\vskip 0.3in
]



\printAffiliationsAndNotice{}  

\begin{abstract}
Dataset distillation aims to synthesize a small number of images per class (IPC) from a large dataset to approximate full dataset training with minimal performance loss. While effective in very small IPC ranges, many distillation methods become less effective, even underperforming random sample selection, as IPC increases. 
 Our examination of state-of-the-art trajectory-matching based distillation methods across various IPC scales reveals that these methods struggle to incorporate the complex, rare features of harder samples into the synthetic dataset even with the increased IPC, resulting in a persistent coverage gap between easy and hard test samples.
Motivated by such observations, we introduce SelMatch, a novel distillation method that effectively scales with IPC. SelMatch uses selection-based initialization and partial updates through trajectory matching to manage the synthetic dataset's desired difficulty level tailored to IPC scales. When tested on CIFAR-10/100 and TinyImageNet, SelMatch consistently outperforms leading selection-only and distillation-only methods across subset ratios from 5\% to 30\%.
\end{abstract}

\section{Introduction}
\label{sec:introduction}

Dataset reduction, essential for data-efficient learning, involves synthesizing or selecting a smaller number of samples from a large dataset, while ensuring that models trained on this reduced dataset still perform comparably or occur minimal performance reduction compared to those trained with the full dataset.  This approach addresses challenges associated with training neural networks on large  datasets, such as high computational costs and memory requirements. 

A prominent technique in this field is dataset distillation, also known as dataset condensation \cite{dd, dc, dsa, dm, tm}. This method distills a large dataset into a smaller, synthetic one.
Data distillation has shown remarkable performance in image classification tasks, especially at extremely small scale, compared to  coreset selection methods \cite{herding,kcenter,craig,gradmatch}. For example, the Matching Training Trajectories (MTT) algorithm \cite{tm} achieves 71.6\% accuracy on a simple ConvNet \cite{convnet} using only 1\% of the CIFAR-10 dataset \cite{cifar}, closely approaching the 84.8\% accuracy of the full dataset. This remarkable efficiency arises from the optimization process, where synthetic samples are optimally learned in a continuous space, instead of being directly selected from the original dataset.

However, recent studies indicate that many dataset distillation methods lose effectiveness and can even underperform compared to random sample selection as the scale of the synthetic dataset or images per class (IPC) increases \cite{dcbench, dq, datm}. This is counterintuitive, considering the greater optimization freedom that distillation provides over discrete sample selection. Specifically, DATM \cite{datm} investigated this phenomenon by analyzing the training trajectories of the state-of-the-art MTT method, noting how the effectiveness of distilled datasets is significantly influenced by the stage of the training trajectories the method focuses on during synthesizing the dataset; particularly, easy patterns learned in early trajectories and hard patterns in later stages distinctly impact MTT's performance across different IPCs. 

We further examine the MTT method across varying IPC, by comparing the coverage of easy versus hard real samples by the synthetic dataset as IPC increases, revealing that the distillation method fails to adequately incorporate the rare features of hard samples to the synthetic dataset even with the increased IPC. This leads to a consistent coverage gap between easy vs. hard test samples, highlighting that the reduced efficacy of dataset distillation methods at larger IPC ranges is partially attributed to their tendency to focus on the simpler, more representative features of the dataset. 
Conversely, as IPC increases, the inclusion of harder, rarer features becomes more crucial for the generalization ability of models trained on the reduced dataset, as demonstrated in data selection studies, both empirically \cite{forgetting, cscore} and theoretically \cite{kolossov2023towards, sorscher2022beyond}. 

Motivated by such observations, we propose a novel method, SelMatch, as a solution for effectively scaling up dataset distillation methods. The key intuition is that as IPC increases, the synthetic dataset should encompass more complex and diverse features of real dataset with suitable difficulty level. The core idea is to manage the desired difficulty level of the synthetic dataset by selection-based initialization and partial updates through trajectory matching. 

$\rhd$ \textbf{Selection-Based Initialization}: To overcome the limitations of the traditional trajectory matching methods overally focusing on easy patterns even with the increased IPC, we propose to initialize the synthetic dataset with real images of \textit{suitable difficulty level} optimized for each IPC. Traditional trajectory matching methods typically initialize the synthetic dataset, with randomly selected samples \cite{dd, dc, dm, tm}, or with easy or representative samples near the class centers \cite{dcbench} to improve the convergence speed of distillation. Our approach is novel in the sense that we initialize the synthetic dataset with a carefully chosen subset, which contains samples of an appropriate difficulty level tailored to the size of the synthetic dataset. This approach ensures that the subsequent distillation process starts with samples of an optimized difficulty level for the specific IPC regime. Experimental results show that selection-based initialization plays an important role in the performance. 

\begin{figure*}[t]
  \begin{center}
      \includegraphics[width=.95\linewidth]{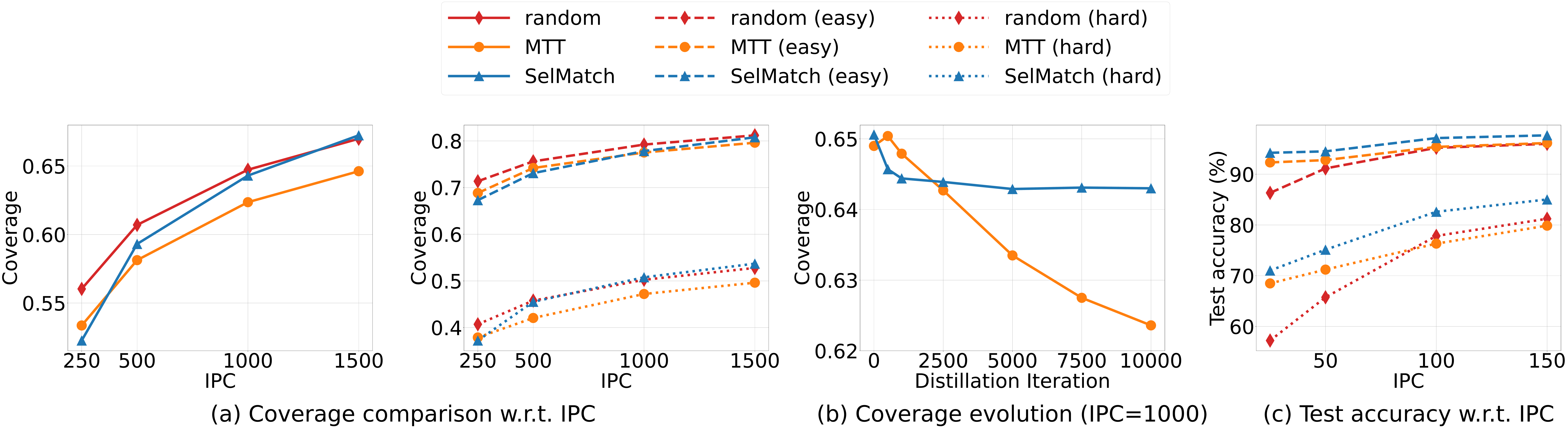}
  \end{center}
  \caption{(a) (left) Overall coverage and (right) coverage of easy vs. hard groups with varying IPC. We observe that coverage by MTT saturates as IPC increases, especially for the hard group. SelMatch (our method) exhibits superior overall coverage, with marked improvements for the hard group. (b) Coverage by MTT decreases rapidly as the distillation proceeds, while that with SelMatch remains stable. (c) Test accuracy on easy vs. hard groups with varying IPC. With MTT, the test accuracy for the hard group eventually aligns with that achieved by random selection as IPC increases. All this findings indicate that traditional MTT overly focuses on synthesizing easy features, leading to saturation in both coverage and test accuracy even with higher IPCs.  In contrast, our method, SelMatch, achieves effective scaling with IPC, enhancing coverage for both easy and hard samples and consequently achieving superior test accuracy.}
  \label{fig:motivation}
\end{figure*}

$\rhd$ \textbf{Partial Updates}: 
In traditional dataset distillation methods, every sample in synthetic dataset is updated during the distillation iterations. 
However, this process keeps reducing the diversity of samples in the synthetic dataset as the distillation iteration increases (as observed in Fig.  \ref{fig:motivation}), since the distillation provides a signal biased toward easy patterns of the full dataset. Thus, to maintain the rare and complex features of hard samples -- essential for model's generalization ability in larger IPC ranges -- we introduce partial updates to the synthetic dataset. The main idea is to keep a fixed portion of the synthetic dataset as unchanged, while updating the rest portion by the distillation signal. 
The ratio of unchanged portion is adjusted according to the IPC. Experimental results show that such partial updates is an important knob for effective scaling of dataset distillation.

We evaluate our method, SelMatch, on the CIFAR-10/100 \cite{cifar} and TinyImageNet \cite{tiny, imagenet} benchmarks, and demonstrate its superiority over state-of-the-art selection-only and distillation-only methods in settings ranging from 5\% to 30\% subset ratios. Remarkably, for CIFAR-100, in the setting with 50 images per class (10\% ratio), SelMatch exhibits 3.5$\%$ increase in test accuracy compared to the leading method.
Our code is publicly available at \href{https://github.com/Yongalls/SelMatch}{https://github.com/Yongalls/SelMatch}.

\section{Related Works}
\label{sec:related}

We begin by reviewing two main approaches in dataset reduction: sample selection and dataset distillation.

\paragraph{Sample Selection} In sample selection, there are two main approaches: optimization-based and score-based selection.
Optimization-based selection aims to identify a small coreset that effectively represents the full dataset's diverse characteristics. For example, Herding \cite{herding} and  K-center \cite{kcenter} select a coreset that approximates the full dataset distributions. 
 Craig \cite{craig} and GradMatch \cite{gradmatch} seek a coreset that minimizes the average gradient difference with the full dataset in neural network training. Although effective in small to intermediate IPCs, these methods often face scalability issues, both computationally and in performance, especially as IPC increases, compared to score-based selection.
Score-based selection assigns value to each instance based on difficulty or influence \cite{infl_func, TracIn} in neural network training. For instance, Forgetting \cite{forgetting} gauges an instance's learning difficulty by counting the number of times it is misclassified after being correctly classified in a previous epoch. C-score \cite{cscore} assesses difficulty as the probability of incorrectly classifying a sample when it is omitted from the training set. These methods prioritize difficult samples, capturing rare and complex features, and outperform optimization-based selection methods, especially in larger IPC scales \cite{EL2N,forgetting}. 
These studies show that, as IPC increases, incorporating harder or rarer features is increasingly important for the improved generalization capability of a model.

\paragraph{Dataset Distillation} Dataset distillation, introduced in \cite{dd}, aims to create a small synthetic set $\mathcal{S}$, so that a model $\theta^\mathcal{S}$ trained on $\mathcal{S}$ achieves good generalization performance, performing well on the full dataset $\mathcal{T}$:
\begin{equation}\nonumber
\mathcal{S^*} = \argmin_\mathcal{S} \mathcal{L}^\mathcal{T}(\theta^\mathcal{S}), \text{ with } \theta^\mathcal{S} = \argmin_\theta \mathcal{L}^\mathcal{S}(\theta)
\end{equation}
Here, $\mathcal{L}^\mathcal{T}$ and $\mathcal{L}^\mathcal{S}$ are losses on $\mathcal{T}$ and $\mathcal{S}$, respectively. To tackle the bi-level optimization's computational complexity and memory demands, existing works have employed two methods: surrogate-based matching and kernel-based approaches.
Surrogate-based matching replaces the complex original objective with simpler proxy tasks. For instance, DC \cite{dc}, DSA \cite{dsa}, and MTT \cite{tm} aim to align the trajectory of model $\theta^\mathcal{S}$, trained on $\mathcal{S}$, with that of the full dataset $\mathcal{T}$ by matching gradients or trajectories. DM \cite{dm} ensures $\mathcal{S}$ and $\mathcal{T}$ have similar distributions in feature space. Kernel-based methods, alternatively, approximate neural network training for $\theta^\mathcal{S}$ using kernel methods and derive a closed-form solution for the inner optimization. Examples include KIP \cite{kip, kip2} using kernel-ridge regression with the Neural Tangent Kernel (NTK) and FrePo \cite{frepo} reducing training costs by focusing only on regression over the last learnable layer. 
However, both surrogate-based matching and kernel-based approaches struggle to scale effectively, either computationally or performance-wise, as IPC increases. 
DC-BENCH \cite{dcbench} reported that these methods underperform compared to random sample selection at higher IPCs. 

Recent research has sought to address the scalability issues of the state-of-the-art MTT methods, focusing on either computational aspects, by reducing memory requirements \cite{tesla}, or performance aspects, by harnessing the training trajectory of the full dataset in later epochs \cite{datm, seqmatch}. In particular, DATM \cite{datm} discovered that aligning with early training trajectories enhances performance in low IPC regimes, while aligning with later trajectories is more beneficial in high IPC regimes. Based on this observation, DATM optimized the trajectory-matching range based on IPCs to adaptively incorporate easier or harder patterns from expert trajectories, thus improving MTT's scalability. While DATM efficiently determines the lower and upper bounds of trajectory-matching range based on the tendency of change in matching loss outside these bounds, explicitly quantifying or searching for the desired difficulty level within training trajectories remains as a challenging task. In contrast, our method, SelMatch, employs selection-based initialization and partial updates through trajectory matching to incorporate complex features of hard samples suited to each IPC. In particular, our approach introduces the novelty of initializing synthetic samples with a tailored difficulty level for each IPC range, a strategy not explored in previous literature on dataset distillation. Furthermore, unlike DATM, which is specifically tailored to enhance MTT, the main components of SelMatch, selection-based initialization and partial updates, have broader applicability across various distillation methods (Appendix \ref{subsec:other_baselines}). We demonstrate the effectiveness of our approach, comparing to leading selection-only or distillation-only methods, including DATM in Sec. \ref{sec:experiments}.

\section{Motivation}\label{sec:motivation}

\begin{figure*}[!t]
  \begin{center}
      \includegraphics[width=0.96\linewidth]{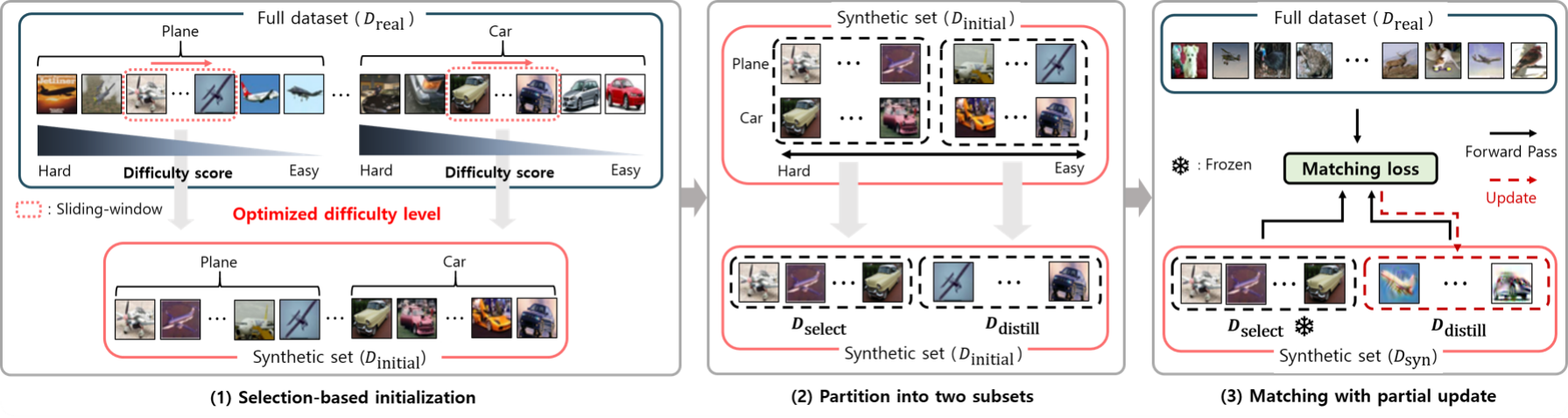}
  \end{center}
  \caption{Illustration of our select-and-match method, SelMatch. Our method comprises two key components: 1) Selection-based initialization: SelMatch employs our sliding-window algorithm to select a subset of a suitable difficulty level, initializing the synthetic dataset $\mathcal{D}_\textrm{syn}$ with this chosen subset; 2) Partial update: SelMatch freezes (1-$\alpha$) fraction of samples ($\mathcal{D}_\textrm{select}$) and update only $\alpha$ fraction of samples ($\mathcal{D}_\textrm{distill}$) while minimizing the matching loss $\mathcal{L}( \mathcal{D}_\textrm{select}\cup \mathcal{D}_\textrm{distill}, \mathcal{D}_\textrm{real})$ to preserve unique features of selected real samples.}
  \label{fig:main}
\end{figure*}

\subsection{Preliminary}
\paragraph{Matching Training Trajectories (MTT)} 
The state-of-the-art dataset distillation method, MTT \cite{tm}, will serve as a baseline to analyze the limitations of traditional dataset distillation methods in large IPC range. 
MTT aims to generate synthetic dataset by matching the training trajectories between the real dataset $\mathcal{D}_\textrm{real}$ and the synthetic dataset $\mathcal{D}_\textrm{syn}$. At each distillation iteration, the synthetic dataset is updated to minimize the matching loss, defined in terms of the training trajectory $\{\theta_t^*\}$  of the real dataset $\mathcal{D}_\textrm{real}$ and that $\{\hat{\theta}_t\}$ of the synthetic dataset $\mathcal{D}_\textrm{syn}$: 
\begin{equation} \label{matching loss}
    \mathcal{L}(\mathcal{D}_\textrm{syn},  \mathcal{D}_\textrm{real})= \frac{\|\hat{\theta}_{t+N} - \theta^*_{t+M}\|^2_2}{\|\theta^*_{t} - \theta^*_{t+M}\|^2_2},
\end{equation}
where $\theta_t^*$ is the model parameter trained on  $\mathcal{D}_\textrm{real}$ at step $t$. Starting from  $\hat{\theta}_{t}=\theta_t^*$, $\hat{\theta}_{t+N}$ refers to the model parameter obtained by training with $\mathcal{D}_\textrm{syn}$ for $N$ steps, while ${\theta}^*_{t+M}$ is the parameter obtained after training on $\mathcal{D}_\textrm{real}$ for $M$ steps.

\subsection{Limitations of Traditional Methods in Larger IPC}\label{sec:coverage}
We first analyze how the patterns of synthetic data, produced by MTT, evolve as the images per class (IPC) increases. 
For a dataset distillation method to remain effective in larger synthetic datasets, the distillation process should keep providing novel and intricate patterns of the real dataset to the synthetic samples as IPC increases.  We initially demonstrate that the trajectory matching methods, though state-of-the-art at low IPC levels, fall short in achieving this objective.

We show this by examining the `coverage' for the real (test) dataset. `Coverage' is defined as the proportion of real samples that fall within a certain radius ($r$) of synthetic samples in the feature space. The radius $r$ is set to the average nearest neighbor distance of the real training samples in the feature space.  A higher coverage indicates that the synthetic dataset captures the diverse features of real samples, enabling the model trained on the synthetic dataset to learn not just the easy but also the complex patterns of the real dataset.

In Figure \ref{fig:motivation}a (left), we show how the coverage changes with increasing IPC for CIFAR-10 dataset. Further, in Figure \ref{fig:motivation}a (right), we analyze this for two groups of samples - the `easy' 50\% and the `hard' 50\% (categorized by Forgetting score \cite{forgetting}, a difficulty measure for real samples). The observation reveals that coverage with MTT does not effectively scale with IPC, consistently falling below that of random selection. Moreover, the coverage for hard sample group is much lower than that of the easy group. This indicates that MTT does not effectively embed the hard and complex data patterns to synthetic samples even with the increased IPC, which can be a reason of MTT's ineffective scaling performance. Our method, SelMatch, exhibits superior overall coverage, with a notable gain in coverage for the hard group, especially as IPC increases. 

An additional important finding is the diminishing coverage of MTT as the number of distillation iterations increases, as shown in Figure \ref{fig:motivation}b. This observation further indicates that traditional distillation methods primarily capture the `easy' patterns over multiple iterations, making the synthesized dataset less diverse as the distillation iterations grow. 
In contrast, with SelMatch, coverage remains stable even with increasing iterations. As shown in Figure \ref{fig:motivation}c, coverage also influences test accuracy. A marked difference in coverage between the easy and hard test sample groups translates into a substantial gap in test accuracy between the two groups. SelMatch, enhancing coverage for both groups, leads to improved test accuracy overall, especially for hard group, as IPC increases.

SelMatch achieves this by combining two key ideas: selection-based initialization and partial updates through trajectory matching. 
These concepts and their implications will be further elaborated in the following section.

\section{Main Method: SelMatch}
\label{sec:method}

Figure \ref{fig:main} illustrates the core idea of our method, SelMatch, which combines selection-based initialization with partial updates through trajectory matching. Traditional trajectory matching methods typically initialize the synthetic dataset, $\mathcal{D}_\textrm{syn}$,
 with a randomly selected subset of the real dataset, $\mathcal{D}_\textrm{real}$, without any specific selection criteria. During each distillation iteration, the entire $\mathcal{D}_\textrm{syn}$ is updated to minimize a matching loss $\mathcal{L}(\mathcal{D}_\textrm{syn}, \mathcal{D}_\textrm{real})$, as defined in \eqref{matching loss}.

In contrast, SelMatch begins by initializing $\mathcal{D}_\textrm{syn}$ with a carefully chosen subset, $\mathcal{D}_\textrm{initial}$, which contains samples of an appropriate difficulty level tailored to the size of the synthetic dataset. Then, during each distillation iteration, SelMatch updates only a specific portion, denoted as $\alpha\in[0,1]$, of $\mathcal{D}_\textrm{syn}$ (referred to as $\mathcal{D}_\textrm{distill}$), while the remaining fraction, $(1-\alpha)$, of the dataset (referred to as $\mathcal{D}_\textrm{select}$) is kept unchanged. This process aims to minimize the same matching loss $\mathcal{L}(\mathcal{D}_\textrm{syn}, \mathcal{D}_\textrm{real})$ in \eqref{matching loss}, but with $\mathcal{D}_\textrm{syn}$ now being a combination of $\mathcal{D}_\textrm{distill}$ and $\mathcal{D}_\textrm{select}$. The rationale and details behind these key components will be elaborated below.

\begin{figure}[t]
  \begin{center}
      \includegraphics[width=0.85\linewidth,height=4.5cm]{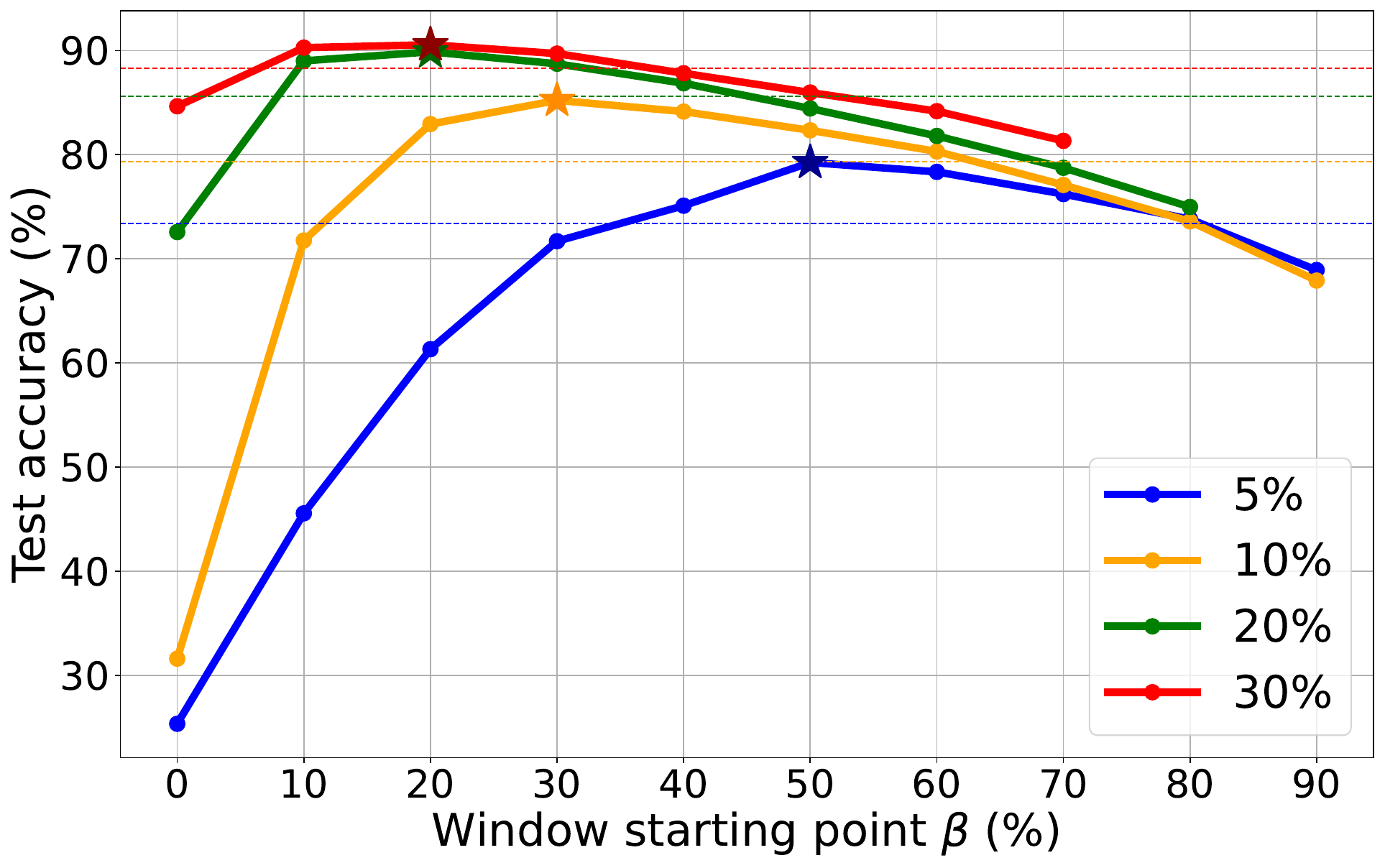}
  \end{center}
  \caption{The result of sliding window experiment on CIFAR-10 with varying subset size (5 to 30\%). Dashed horizontal lines indicate the accuracy of the models trained by randomly selected subsets of the corresponding size. Solid lines indicate the accuracy of the models trained by a window subset of samples ordered by their difficulty scores (from hardest to easiest by c-score \cite{cscore}) with varying window starting point $\beta$\%. }
  \label{fig:window_selection}
\end{figure}
\paragraph{Selection-Based Initialization: Sliding Window Alg.}

An important observation from Fig. \ref{fig:motivation} is that the traditional trajectory matching methods tend to focus on easy and representative patterns of the full dataset rather than complex data patterns, resulting in less effective scaling in larger IPC settings. To overcome this, we propose initializing the synthetic dataset, $\mathcal{D}_\textrm{syn}$, with a carefully selected difficulty level that includes more complex patterns from the real dataset as IPC increases. The challenge then becomes how to select a subset of the real dataset, $\mathcal{D}_\textrm{real}$, with an appropriate level of complexity, considering the size of $\mathcal{D}_\textrm{syn}$.

To address this, we design a sliding window algorithm. We arrange the training samples in descending order of difficulty (most difficult to easiest) based on pre-calculated difficulty scores \cite{cscore, forgetting}. We then assess window subsets of these samples by comparing the test accuracy achieved when a model is trained on each window subset while varying its starting point. For a given threshold $\beta\in[0,100]\%$, after excluding the top $\beta$\% of the hardest samples, the window subset includes samples from the $[\beta, \beta+r]$\% range, where the subset portion is $r=(|\mathcal{D}_\textrm{syn}|/|\mathcal{D}_\textrm{real}|)\times 100\%$ and $|\mathcal{D}_\textrm{syn}|$ equals IPC times the number of classes.  Here, we make sure that each window subset includes equal number of samples from each class.

As depicted in Figure \ref{fig:window_selection}, the starting point of the window, corresponding to the level of difficulty, significantly influences the generalization ability of the model, as measured by test accuracy. Particularly for smaller windows (5-10\% range), we observe up to a 40\% deviation in test accuracy according to where the window starts. Moreover, the best window subsets, achieving the highest test accuracy, tend to include more difficult samples (smaller $\beta$) as the subset size increases, which aligns with the intuition that incorporating complex patterns from the real dataset enhances the model's generalization capability as IPC grows.

Building on this observation, we set the initialization of $\mathcal{D}_\textrm{syn}$ to $\mathcal{D}_\textrm{initial}$, where $\mathcal{D}_\textrm{initial}$ is the best-performing window subset identified by the sliding window algorithm for the given $\mathcal{D}_\textrm{syn}$ size. This approach ensures that the subsequent distillation process starts with images of an optimized difficulty level for the specific IPC regime.

\paragraph{Partial Updates} 

After initializing the synthetic dataset $\mathcal{D}_\textrm{syn}$ with the best window subset $\mathcal{D}_\textrm{initial}$, chosen from the sliding window algorithm, our next goal is to update $\mathcal{D}_\textrm{syn}$ through dataset distillation to efficiently embed the information from the entire real dataset $\mathcal{D}_\textrm{real}$. Traditionally, the matching training trajectories (MTT) algorithm updates  all the samples in $\mathcal{D}_\textrm{syn}$, by backpropagating through $N$ model updates, to minimize the matching loss \eqref{matching loss}. However, as shown in Fig. \ref{fig:motivation}b, this approach tends to favor simpler patterns in the dataset, leading to a reduction in coverage over successive distillation iterations. Thus, to counter this and to maintain the unique and complex features of some real samples -- essential for model's generalization ability in larger IPC ranges -- we introduce partial updates to $\mathcal{D}_\textrm{syn}$.

We partition the initial synthetic dataset $\mathcal{D}_\textrm{syn}=\mathcal{D}_\textrm{initial}$ into two subsets $\mathcal{D}_\textrm{select}$ and $\mathcal{D}_\textrm{distill}$ based on each sample's difficulty score. The subset $\mathcal{D}_\textrm{select}$ contains $(1-\alpha) \times |\mathcal{D}_\textrm{syn}|$ samples with higher difficulty, while the remaining $\alpha$ fraction of samples are assigned to $\mathcal{D}_\textrm{distill}$, where $\alpha\in[0,1]$ is a hyperparameter adjusted according to the IPC. 

During distillation iterations, we keep $\mathcal{D}_\textrm{select}$ unchanged and update only the $\mathcal{D}_\textrm{distill}$ subset. The update aims to minimize the matching loss between the entire $\mathcal{D}_\textrm{syn}=\mathcal{D}_\textrm{select}\cup \mathcal{D}_\textrm{distill}$ and $\mathcal{D}_\textrm{real}$, i.e., 
\beq \label{eq:partial_update}
\mathcal{L}( \mathcal{D}_\textrm{select}\cup \mathcal{D}_\textrm{distill}, \mathcal{D}_\textrm{real}),
\eeq
rather than minimizing $\mathcal{L}(\mathcal{D}_\textrm{distill}, \mathcal{D}_\textrm{real})$, the loss considering only the updated portion of $\mathcal{D}_\textrm{syn}$. This  strategy encourages $\mathcal{D}_\textrm{distill}$ to condense the knowledge not present in $\mathcal{D}_\textrm{select}$, thereby enriching the overall information within $\mathcal{D}_\textrm{syn}$. 

\paragraph{Combined Augmentation}
After creating the synthetic dataset $\mathcal{D}_\textrm{syn}$, we assess its effectiveness by training a randomly initialized neural network using this dataset.  Typically, previous distillation methods have employed Differentiable Siamese Augmentation (DSA) \cite{dsa} to evaluate synthetic datasets. This approach, involving more complex augmentation techniques than the simpler methods (like random cropping and horizontal flipping) commonly used for real datasets \cite{krizhevsky2012imagenet}, has been shown to yield better results for synthetic data, as noted in empirical studies \cite{dcbench}. This improved performance may arise because synthetic datasets predominantly capture easier patterns, making them more suitable for the stronger augmentation by DSA.

However, applying DSA across our entire synthetic dataset $\mathcal{D}_\textrm{syn}$ might not be ideal, especially considering the presence of the subset $\mathcal{D}_\textrm{select}$, which contains diifficult samples. To address this, we propose a combined augmentation strategy tailored to our synthetic dataset. Specifically, we apply DSA to the distilled portion, $\mathcal{D}_\textrm{distill}$, and use the simpler, more traditional augmentation techniques for the selected, more complex subset $\mathcal{D}_\textrm{select}$. This combined approach aims to leverage the strengths of both augmentation methods to enhance the overall performance of the synthetic dataset.

Putting all together, SelMatch is summarized in Alg. \ref{alg:main}.

\section{Experimental Results}
\label{sec:experiments}


\begin{table*}[t]
\centering\scriptsize
\begin{adjustbox}{max width=\linewidth}
\begin{threeparttable}
\caption{Performance of SelMatch compared to other baselines on CIFAR-10, CIFAR-100 and Tiny ImageNet. We highlight the best result across all methods in bold and underline the best result among selection-only baselines. \DAG denotes the use of distilled dataset provided by the original paper without reproduction. For all subset ratios, we set the number of training steps to be 25\% of that with full dataset for 200 epochs.  For the full dataset, we also report the test accuracy measured by training for the full 200 epochs in bracket.}
\label{tab:main_result}
\vskip 0.15in
\centering
\begin{tabular}{c|cccc|cccc|cc}
\toprule
Dataset & \multicolumn{4}{c|}{CIFAR-10} & \multicolumn{4}{c|}{CIFAR-100} & \multicolumn{2}{c}{Tiny ImageNet} \\
IPC & 250 & 500 & 1000 & 1500 & 25 & 50 & 100 & 150 & 50 & 100 \\
Ratio & $5\%$ & $10\%$ & $20\%$ & $30\%$ & $5\%$ & $10\%$ & $20\%$ & $30\%$ & $10\%$ & $20\%$ \\
\midrule
Random & 73.4$\pm$1.5 & 79.3$\pm$0.3 & 85.6$\pm$0.4 & 88.3$\pm$0.2 & 35.8$\pm$0.6 & 40.7$\pm$1.0 & 53.2$\pm$0.9 & 60.3$\pm$1.3 & 30.1$\pm$0.6 & 40.1$\pm$0.4 \\
Forgetting & 30.7$\pm$0.3 & 41.5$\pm$0.7 & 68.4$\pm$1.6 & 83.5$\pm$1.8 & 9.5$\pm$0.3 & 13.2$\pm$0.6 & 27.0$\pm$1.1 & 42.3$\pm$1.0 & 5.7$\pm$0.1 & 12.4$\pm$0.2 \\
Glister & 46.6$\pm$1.3 & 56.6$\pm$0.5 & 79.0$\pm$0.7 & 85.0$\pm$0.9 & 21.7$\pm$0.8 & 26.7$\pm$1.3 & 39.9$\pm$1.4 & 52.1$\pm$1.3 & 22.6$\pm$0.5 & 34.0$\pm$0.3 \\
Oracle window & \underline{79.3$\pm$0.7} & \underline{85.2$\pm$0.1} & \underline{89.9$\pm$0.5} & \underline{90.6$\pm$0.3} & \underline{43.2$\pm$1.8} & \underline{50.0$\pm$0.8} & \underline{59.2$\pm$0.8} & \underline{64.7$\pm$0.5} & \underline{42.5$\pm$0.3} & \underline{49.2$\pm$0.3} \\
\midrule
DSA\tnote{1} & 74.7$\pm$1.5 & 78.7$\pm$0.7 & 84.8$\pm$0.5 & - & 38.4$\pm$0.4 & 43.6$\pm$0.7 & - & - & 27.8$\pm$1.4\DAG & - \\
DM\tnote{1} & 75.3$\pm$1.4 & 79.1$\pm$0.6 & 85.6$\pm$0.5 & - & 37.5$\pm$0.6 & 42.6$\pm$0.5 & - & - & 31.0$\pm$0.6\DAG & - \\
MTT\tnote{2} & 80.7$\pm$0.4 & 82.2$\pm$0.4 & 86.1$\pm$0.3 & 88.6$\pm$0.2 & 49.9$\pm$0.7 & 51.3$\pm$0.4 & 58.7$\pm$0.6 & 63.1$\pm$0.3 & 40.3$\pm$0.3 & 44.2$\pm$0.5 \\
FTD\tnote{3,4} & 78.8$\pm$0.2 & 80.0$\pm$0.7 & 85.6$\pm$0.2 & 87.8$\pm$0.4 & 47.4$\pm$0.2 & 49.0$\pm$0.2 & 56.2$\pm$0.3 & 61.0$\pm$0.5 & 30.2$\pm$0.3 & 36.9$\pm$0.6 \\
DATM\tnote{3} & - & 84.8$\pm$0.3\DAG & 87.6$\pm$0.3\DAG & - & - & 51.0$\pm$0.5\DAG & 61.5$\pm$0.3\DAG & - & 42.2$\pm$0.2\DAG & - \\
\midrule
SelMatch(ours) & \textbf{82.8}$\pm$\textbf{0.2} & \textbf{85.9}$\pm$\textbf{0.2} & \textbf{90.4}$\pm$\textbf{0.2} & \textbf{91.3}$\pm$\textbf{0.2} & \textbf{50.9}$\pm$\textbf{0.3} & \textbf{54.5}$\pm$\textbf{0.6} & \textbf{62.4}$\pm$\textbf{0.6} & \textbf{67.4}$\pm$\textbf{0.2} & \textbf{44.7}$\pm$\textbf{0.2} & \textbf{50.4}$\pm$\textbf{0.2} \\
\midrule
Full Dataset & \multicolumn{4}{c|}{93.2$\pm$0.3 (95.0$\pm$0.2)} & \multicolumn{4}{c|}{73.9$\pm$0.2 (76.5$\pm$0.7)} & \multicolumn{2}{c}{58.5$\pm$0.3 (61.1$\pm$0.2)} \\
\bottomrule
\end{tabular}
\vskip -0.1in
\begin{tablenotes}
    \item [1] We reproduce DSA and DM only on the scalable regime: CIFAR-10 with IPC ranging from 250 to 1000, and CIFAR-100 with IPC ranging from 25 to 50, following dc-benchmark \cite{dcbench}.
    \item [2] We reproduce MTT with ConvNet-BN and with larger max start epoch $T^+$ than used in the original paper. 
    \item [3] ZCA whitening is used in FTD and DATM.
    \item [4] EMA (exponential moving average) is used in FTD.
\end{tablenotes}
\end{threeparttable}
\end{adjustbox}
\end{table*}

\subsection{Experiment Setup}
\label{subsec:experiment_setup}

We evaluate the performance of our method, SelMatch, on various datasets including CIFAR-10/100, and Tiny ImageNet. Considering that SelMatch is a combined mechanism of selection and distillation method, we compare our method with both selection-only and distillation-only baselines. For the selection-only baselines, we incorporate two sample selection methods, Forgetting \cite{forgetting}, representing difficulty score-based selection, and Glister \cite{glister}, representing optimization-based coreset selection. Additionally, we present the result of oracle-window selection, which denotes the optimal subset $\mathcal{D}_\textrm{initial}$ obtained by the sliding window algorithm (Fig. \ref{fig:window_selection}). For distillation-only baselines, we include DSA \cite{dsa}, DM \cite{dm}, MTT \cite{tm}, FTD \cite{ftd}, and DATM \cite{datm}. Due to scalability issues, kernel-based methods are excluded from our experiment. \\

It is noteworthy that previous distillation methods use ConvNet architecture \cite{convnet, dc} with Instance Normalization \cite{instancenorm} (denoted as ConvNet-IN) for both distillation and evaluation instead of Batch Normalization (ConvNet-BN) \cite{batchnorm}. In contrast, we employ ConvNet-BN for the distillation and ResNet18 architecture \cite{resnet} for the evaluation. For a fair comparison, we evaluate all methods with ResNet18 architecture and DSA augmentation (combined augmentation for SelMatch). We report the mean and standard deviation of test accuracy by training 5 randomly initialized networks with the reduced dataset. Note that, we set the number of training steps to be 25\% of that required for training with the full dataset over 200 epochs. Specifically, we train networks for 1,000, 500, 250, and 167 epochs when evaluating reduced dataset of $5\%$, $10\%$, $20\%$, and $30\%$ subset ratios, respectively. More details of experiment setup are presented in Appendix \ref{sec:implement_detail}.

\subsection{Main Results}

\paragraph{CIFAR and Tiny ImageNet}We compare SelMatch against the existing sample selection (Random, Forgetting, Glister) and dataset distillation (DSA, DM, MTT, FTD, DATM) baselines for CIFAR-10/100 and Tiny ImageNet across 5\% to 30\% selection ratios. Oracle window is the result obtained by using the window subset $\mathcal{D}_\textrm{initial}$ selected from the sliding window algorithm (Fig. \ref{fig:window_selection}) but without the subsequent dataset distillation. 
As shown in Table \ref{tab:main_result}, the performance gains from previous distillation methods saturate rapidly, or even fall behind random selection, as IPC increases. Another interesting observation is that oracle window selection outperforms all other baselines except SelMatch, in every tested ratio on CIFAR-10 and Tiny ImageNet. This result indicates that choosing the subset of a desirable difficulty level tailored to each IPC scale is important in developing effectively scalable dataset reduction methods. Previous distillation methods have not effectively achieved this. Through the optimized selection-based initialization and partial updates, our method establishes the new state-of-the-art performance in all settings. Particularly, we achieve a performance gain of $3.5\%$ compared to the next best method on CIFAR-100 with IPC=50 (subset ratio 10\%). 

\paragraph{Cross-Architecture Generalization}

One of the essential characteristics of a distilled dataset is its ability to generalize across different, unseen architectures. In fact, we have already demonstrated SelMatch's generalization capability by conducting distillation on the ConvNet architecture and evaluating it on the ResNet18 architecture. Nevertheless, we further evaluate the cross-architecture generalizability by considering additional network architectures, such as VGG \cite{vgg}. As presented in Table \ref{tab:cross_arch}, our method exhibits competitive performance on unseen architectures. We notice that our method shows lower performance than DATM on simpler architectures (ConvNet and VGG-11) but still outperforms MTT. This outcome is a natural consequence, as our distilled dataset contains complex features that are challenging to learn with smaller networks like ConvNet.

\begin{table}[t]
\caption{Cross-architecture experiment results on CIFAR-10 with IPC=1000. Our method generalizes well on unseen network architectures, especially on large networks.}
\label{tab:cross_arch}
\vskip 0.15in
\begin{center}
\begin{scriptsize}
\resizebox{\linewidth}{!}
{%
\begin{tabular}{c|cccccc}
\toprule
Method & Conv & Res-18 & Res-34 & Res-50 & VGG-11 & VGG-16\\
\midrule
Random & 79.8 & 85.6 & 85.8 & 85.9 & 81.9 & 84.8 \\
MTT & 82.3 & 86.1 & 85.7 & 85.8 & 83.9 & 85.6 \\
DATM & \textbf{84.8} & 87.6 & 87.2 & 86.9 & \textbf{86.3} & 86.3 \\
SelMatch (ours) & 83.5 & \textbf{90.4} & \textbf{89.9} & \textbf{89.3} & 86.0 & \textbf{88.2} \\
\bottomrule
\end{tabular}
}
\end{scriptsize}
\end{center}
\vskip -0.1in
\end{table}

\subsection{Ablation Study and Further Analysis}
\label{subsec:ablation_study}

\begin{table}[t]
\caption{Ablation study on key components of our method on CIFAR-100 with IPC=100. Note that the top row indicates the baseline (MTT) and the last row indicates our method, SelMatch.}
\label{tab:ablation_components}
\vskip 0.15in
\begin{center}
\begin{scriptsize}
\begin{tabular}{ccc|c}
\toprule
Select Init & Partial Update & Combined Augment & Acc \\
\midrule
\xmark & \xmark & \xmark & 58.7 \\
\checkmark & \xmark & \xmark & 61.0 \\
\checkmark & \checkmark & \xmark & 61.5 \\
\checkmark & \checkmark & \checkmark & 62.4 \\
\bottomrule
\end{tabular}
\end{scriptsize}
\end{center}
\vskip -0.1in
\end{table}

\begin{table}[t]
\caption{Ablation on the way of using selection and distillation on CIFAR-100 with varying IPC. }
\label{tab:ablation_merge}
\vskip 0.15in
\begin{center}
\begin{scriptsize}
\begin{tabular}{l|cccc}
\toprule
IPC & 25 & 50 & 100 & 150 \\
\midrule
Selection-only (Oracle window) & 43.2 & 50.0 & 59.2 & 64.7 \\
Distillation-only (MTT) & 49.9 & 51.3 & 58.7 & 63.1 \\
Merge & 51.1 & 53.1 & 61.7 & 67.0 \\
SelMatch (ours) & 50.9 & 54.5 & 62.4 & 67.4 \\
\bottomrule
\end{tabular}
\end{scriptsize}
\end{center}
\vskip -0.1in
\end{table}
\begin{figure*}[t]
\centering
    \subfigure[Ablation on $\alpha$\label{fig:ablation_alpha}]
    {\includegraphics[width=0.24\linewidth]{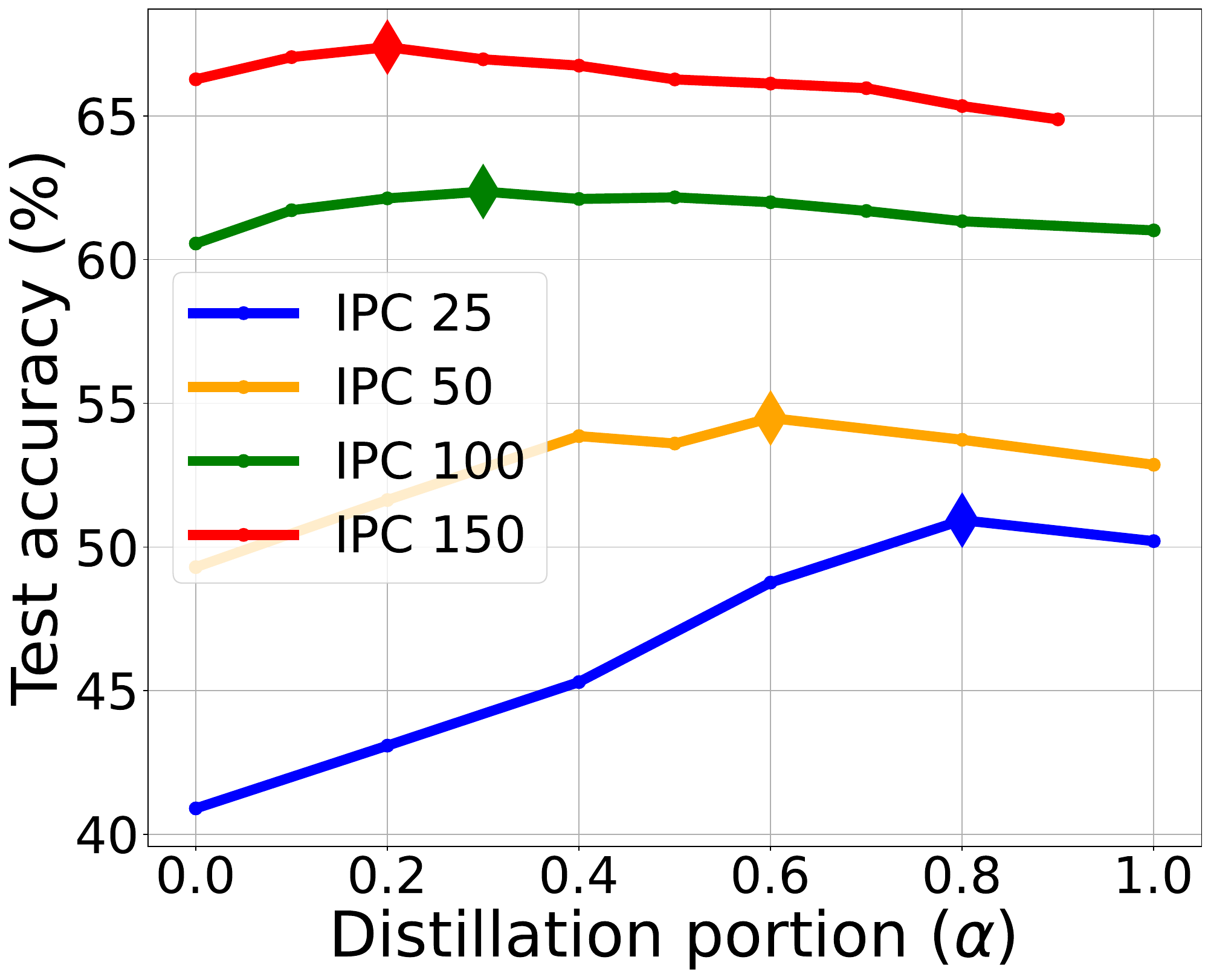}}
    \subfigure[Ablation on augmentation\label{fig:ablation_aug_ipc50}]{\includegraphics[width=0.24\linewidth]{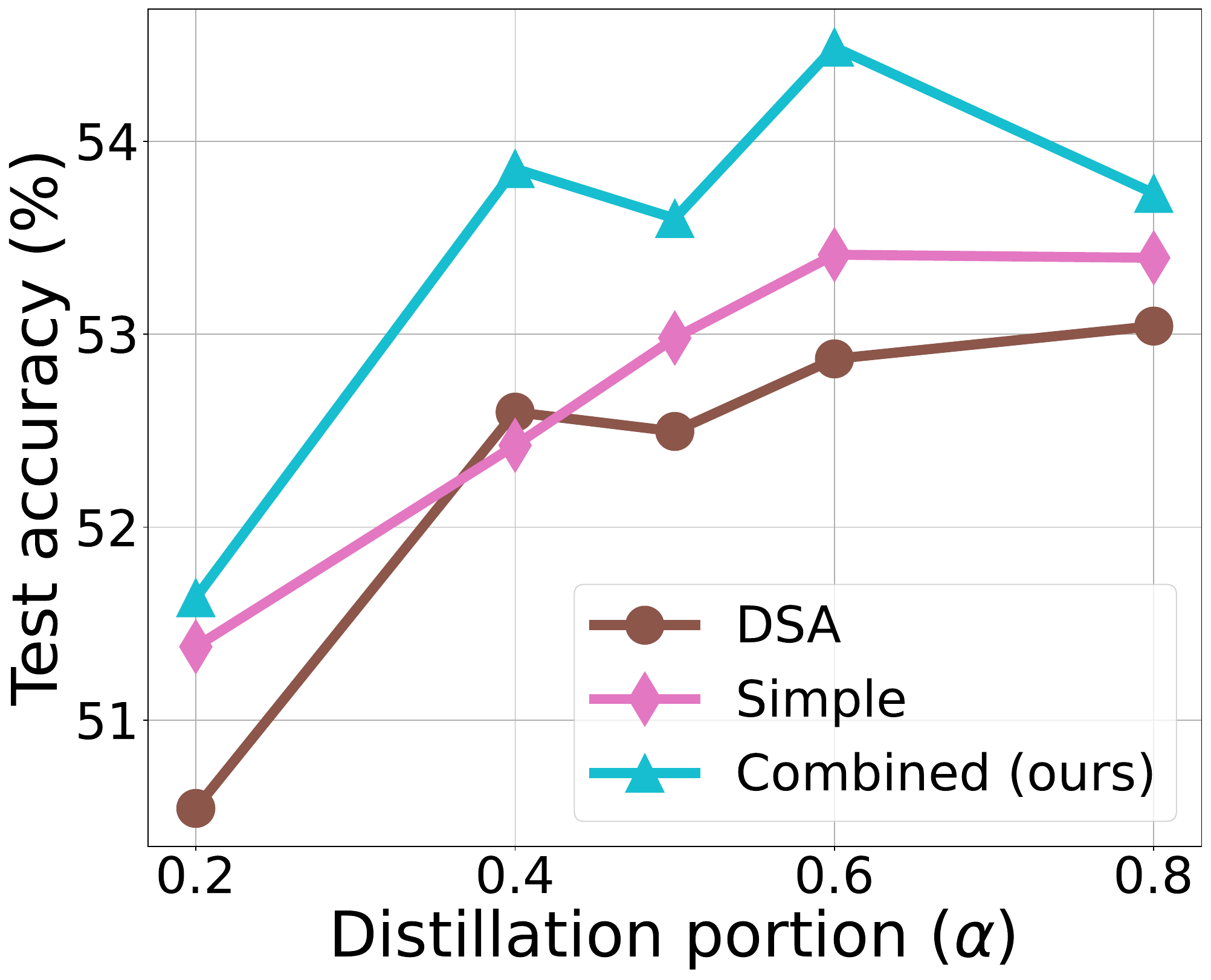}}
    \subfigure[Ablation on BN\label{fig:ablation_BN}]{\includegraphics[width=0.255\linewidth]{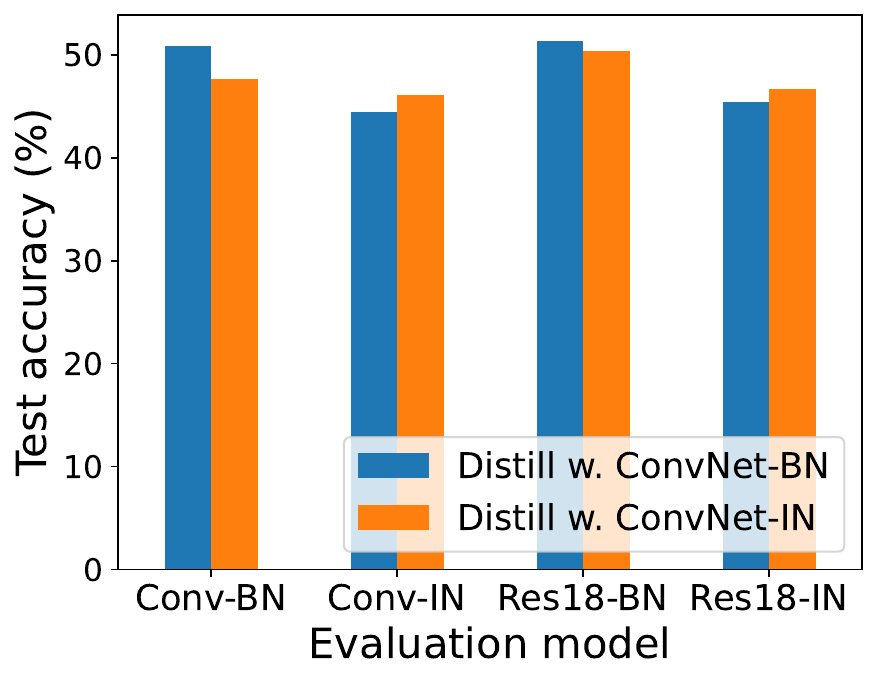}}
    \subfigure[Ablation on $T^+$\label{fig:ablation_T}]{\includegraphics[width=0.24\linewidth]{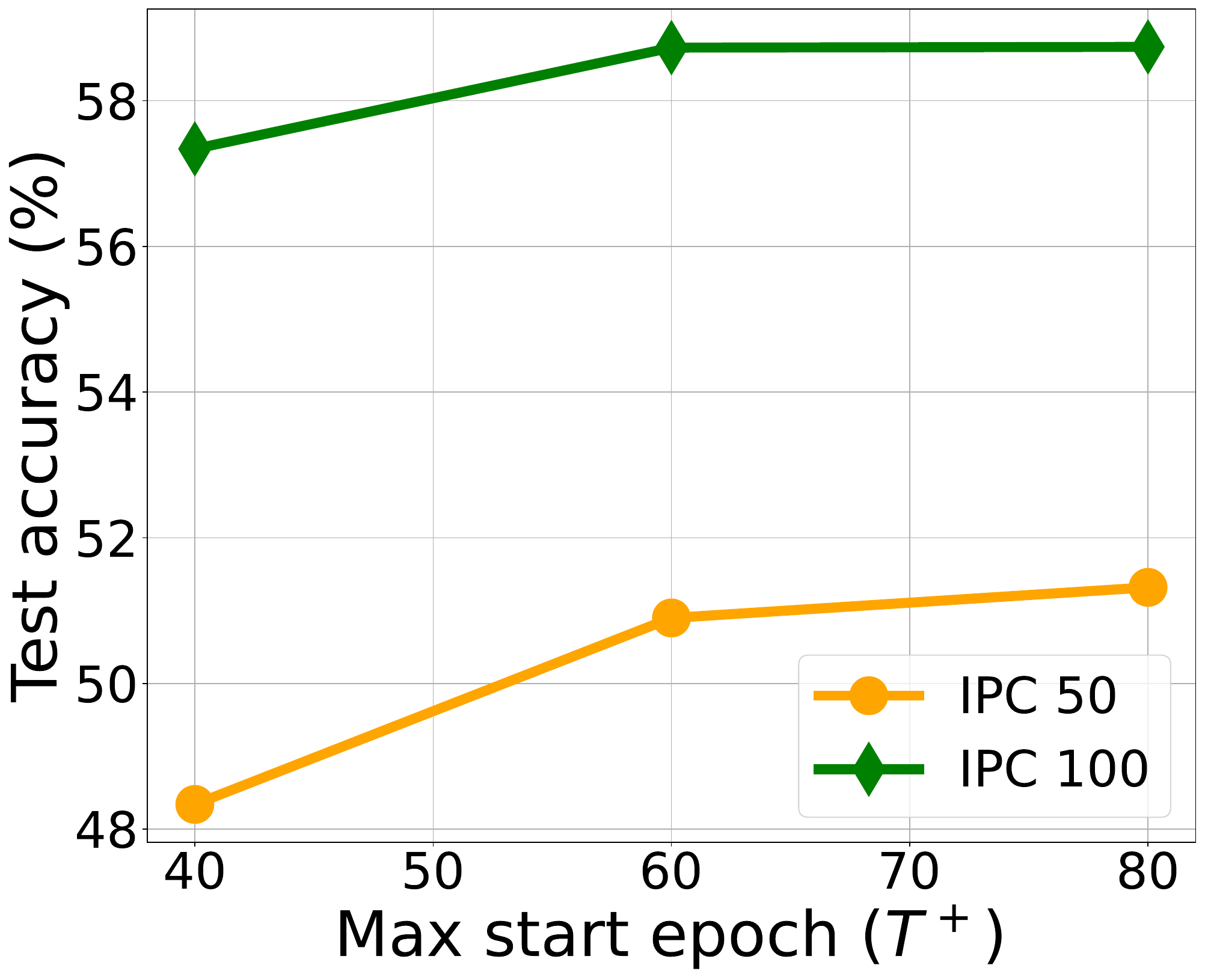}}
    
    \caption{(a) Ablation on distillation portion $\alpha$ in synthetic dataset for CIFAR-100 with varying IPC. Optimal $\alpha$ tends to decrease as IPC increases. (b) Ablation on augmentation strategy on CIFAR-100 with IPC 50. The result shows the effectiveness of our combined augmentation technique. (c) Ablation on batch normalization on CIFAR-100 with IPC=50. Employing Batch Normalization for both distillation and evaluation exhibits the best performance. (d) Ablation on max start epoch $T^+$ on CIFAR-100 with IPC=50, 100. The result indicates that utilizing later epochs enhances performance in large IPC regime.
    }
    \label{fig:ablation_alpha_aug}
\end{figure*}

\paragraph{Key Components} We conduct an ablation study to analyze the effect of three key components of SelMatch, described in Section \ref{sec:method}. Specifically, we isolate and measure the effects of 1) selection-based initialization, 2) partial updates, and 3) combined augmentation. Table \ref{tab:ablation_components} presents the ablation results on CIFAR-100 with IPC=100. As shown in the table, all components of our method contribute to performance improvement. Especially, we observe that selection-based initialization leads to a substantial performance boost (+2.3\% compared to MTT), demonstrating the critical significance of carefully initializing the synthetic dataset. Furthermore, partial updates yield additional performance improvements, with the combined augmentation amplifying the effectiveness of partial updates.

\begin{figure*}[t]
\centering
    \subfigure[T-SNE visualization\label{fig:tsne}]
    {\includegraphics[width=0.64\linewidth]{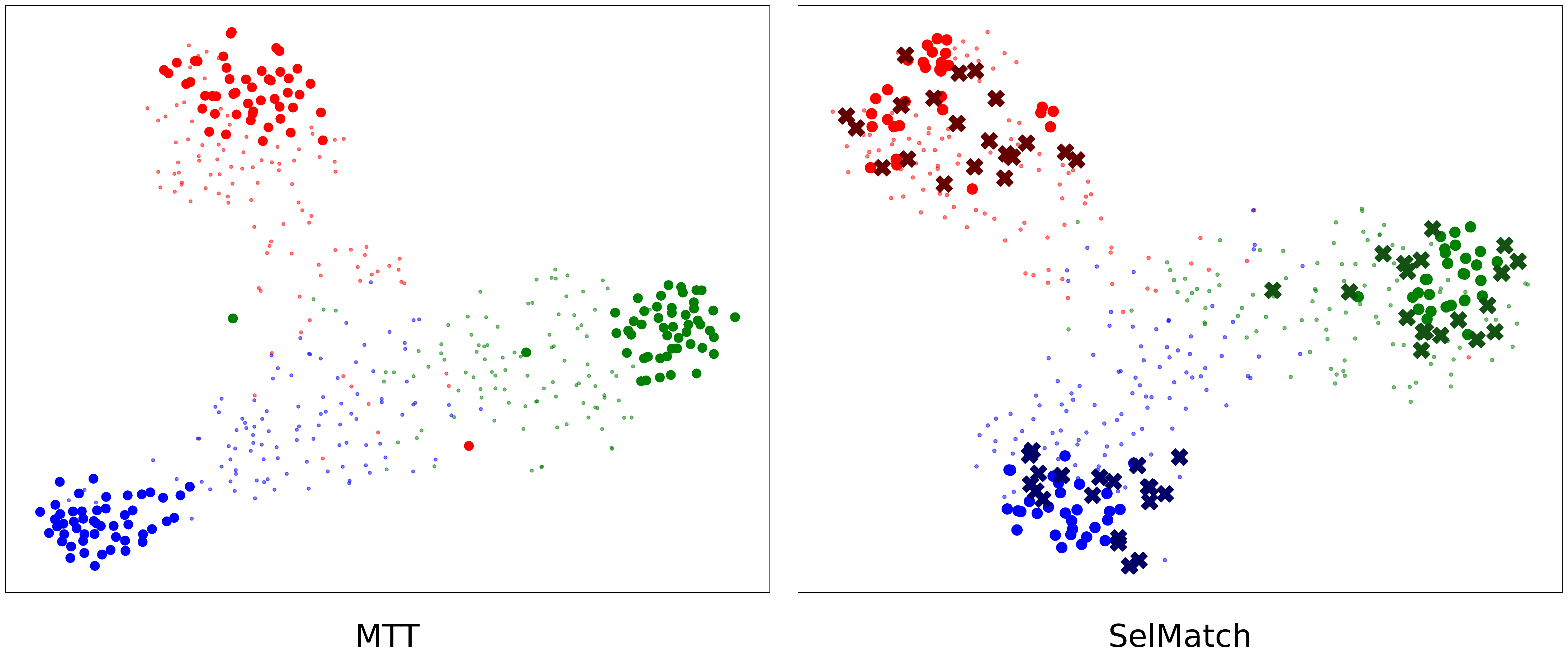}}
    \subfigure[Gradient norm\label{fig:gradient_norm}]
    {\includegraphics[width=0.31\linewidth]{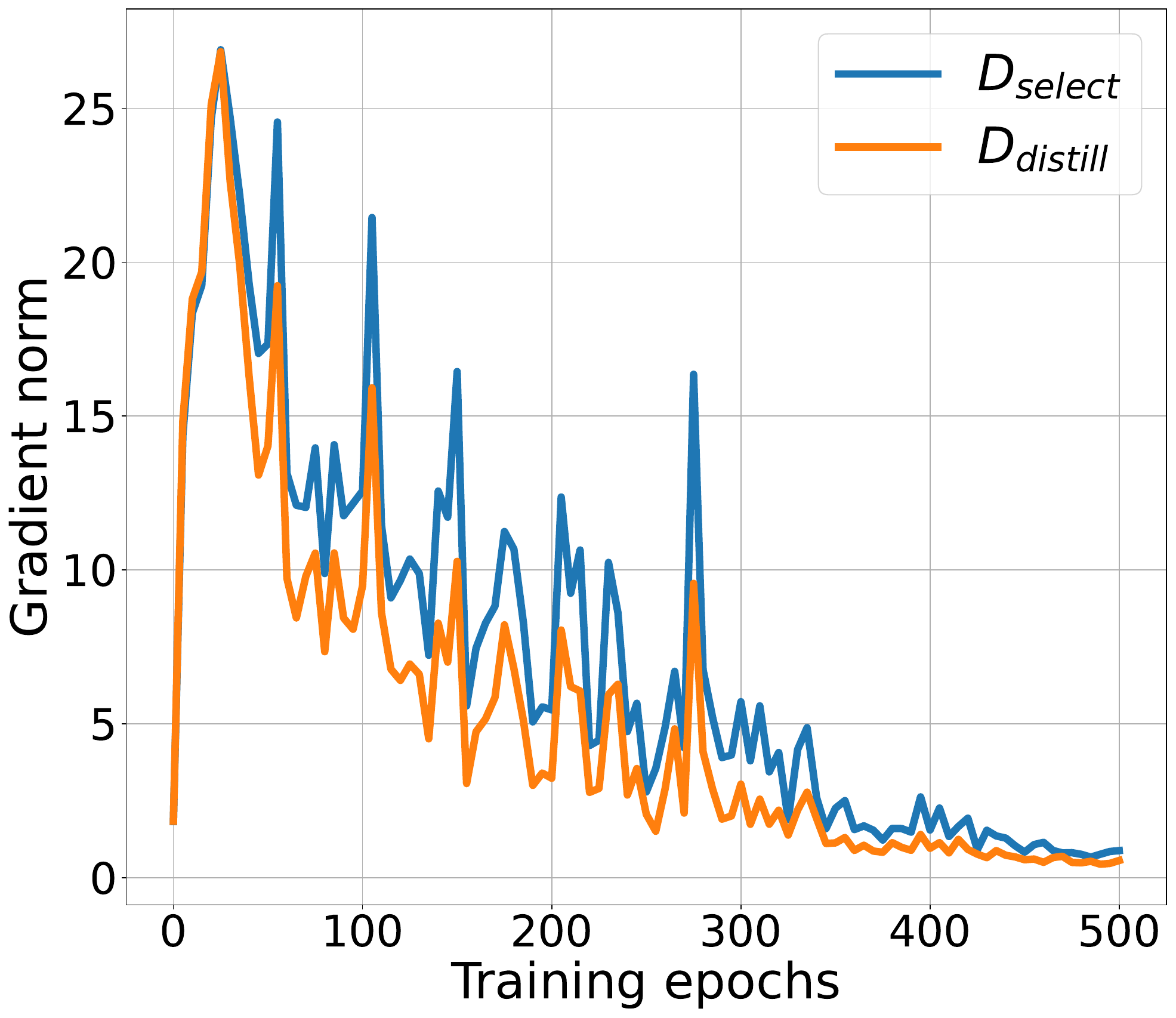}}
    \caption{Analysis of SelMatch on CIFAR-100 with IPC=50. (a) T-SNE visualization of \textit{(left)} MTT and \textit{(right)} SelMatch. Small red, green, blue points represent real samples (test set) of the first three classes of CIFAR-100. Large circles indicate samples in the synthetic dataset. For SelMatch, unaltered samples ($\mathcal{D}_\textrm{select}$) are denoted as `X' marker with darker colors. We observe that samples in $\mathcal{D}_\textrm{select}$ are located closer to the decision boundary compared to $\mathcal{D}_\textrm{distill}$. (b) Evolution of $\ell_2$ norm of network gradients on $\mathcal{D}_\textrm{select}$ and $\mathcal{D}_\textrm{distill}$. The gradient norm on $\mathcal{D}_\textrm{select}$ is larger than $\mathcal{D}_\textrm{distill}$. Note that network is trained on entire synthetic set $\mathcal{D}_\textrm{syn} = \mathcal{D}_\textrm{select} \cup \mathcal{D}_\textrm{distill}$.
    }
    \label{fig:analysis}
\end{figure*}
\paragraph{Selection and Distillation}  
SelMatch updates only a specific portion, denoted as $\alpha\in[0,1]$, of $\mathcal{D}_\textrm{syn}$ (referred to as $\mathcal{D}_\textrm{distill}$), while the remaining fraction, $(1-\alpha)$, of the dataset (referred to as $\mathcal{D}_\textrm{select}$) is kept unchanged from initialization. This process aims to minimize the matching loss $\mathcal{L}(\mathcal{D}_\textrm{distill}\cup\mathcal{D}_\textrm{select}, \mathcal{D}_\textrm{real})$.
A less sophisticated approach is simply using $\mathcal{L}(\mathcal{D}_\textrm{distill}, \mathcal{D}_\textrm{real})$ and merge $\mathcal{D}_\textrm{select}$ and $\mathcal{D}_\textrm{distill}$ after the distillation. We call this method ``Merge". Table \ref{tab:ablation_merge} compares the performances of oracle-window (selection only), MTT (distillation only), Merge (simply merging selected samples with distilled dataset) and our SelMatch. We can observe that simply merging selected samples with distilled dataset (initialized with the optimized $\mathcal{D}_\textrm{initial}$) already brings a significant gain than the traditional MTT. SelMatch brings additional improvement than the simple Merge. 

\paragraph{Distillation portion $\alpha$} Figure \ref{fig:ablation_alpha} shows the effect of distillation portion $\alpha$ in the synthetic dataset for different IPC scales. The result reveals that the optimal value of $\alpha$ decreases as IPC increases, i.e., updating a smaller portion of the synthetic dataset by distillation results in better performance for larger IPC scales. This observation aligns with our intuition that maintaining complex and unique samples has a greater benefit in larger IPC regimes. 

\paragraph{Combined Augmentation} We also study the impact of our combined augmentation in Figure \ref{fig:ablation_aug_ipc50}. We observe that the combined augmentation outperforms both DSA-only and simple-only augmentations. This result demonstrates that our proposed augmentation strategy can efficiently leverage the benefits of combining $\mathcal{D}_\textrm{select}$ and $\mathcal{D}_\textrm{distill}$.

\paragraph{Batch Normalization} Previous distillation works often use instance normalization (IN) instead of  batch normalization (BN), since inaccurate batch statistics of extremely small synthetic dataset can hamper the performance. However, since we target on larger synthetic dataset, we employ batch normalization for SelMatch and reproduction of MTT. To explore the effect of batch normalization, we perform an ablation on CIFAR-100 with IPC=50, both for distillation and evaluation steps by applying IN or BN, as shown in Figure \ref{fig:ablation_BN}. The result reveals that using batch normalization both for distillation and evaluation yields the best result.

\paragraph{Max Start Epoch $T^+$} We perform an ablation study on max start epoch $T^+$ of MTT where the start epoch $t$ in the matching loss \eqref{matching loss} is randomly chosen from $t\leq T^+$. As shown in Fig. \ref{fig:ablation_T}, we observe that increasing $T^+$ improves the performance in large IPC settings. This result indicates that leveraging information of later training epochs is beneficial for larger synthetic dataset. In Table 1, we reproduced MTT with $T^+=80$ and used the same $T^+$ for SelMatch for all cases, while the original paper used $T^+\leq40$.

\paragraph{Further Analysis}

    

The core idea behind SelMatch is to synthesize representative samples while preserving complex and rare features through partial updates. To validate that our method actually works as intended, we investigate the \textit{typicality} of updated subset $\mathcal{D}_\textrm{distill}$ and unaltered subset $\mathcal{D}_\textrm{select}$. In Figure \ref{fig:tsne}, we depict the data distribution of MTT and SelMatch. We train networks on each synthetic dataset to extract features and visualize these features using T-SNE \cite{tsne}. The visualization reveals that samples in $\mathcal{D}_\textrm{select}$ (with `X' marker) are positioned closer to the decision boundary than samples in $\mathcal{D}_\textrm{distill}$. Consequently, SelMatch demonstrates the ability to incorporate more diverse and hard samples in the real distribution than MTT. Additionally, we measure the network's gradient norm on $\mathcal{D}_\textrm{distill}$ and $\mathcal{D}_\textrm{select}$ throughout the training. As shown in Figure \ref{fig:gradient_norm}, we observe that $\mathcal{D}_\textrm{select}$ generates a larger gradient norm, indicating that $\mathcal{D}_\textrm{select}$ contains more complex and intricate patterns compared to $\mathcal{D}_\textrm{distill}$.


\section{Discussions}

To address the limitations of current distillation methods, which struggle to scale effectively with increasing IPC,  we introduced SelMatch, a novel distillation method that combines selection-based initialization with partial updates through trajectory matching. Our approach ensures that the distillation process begins with images at an optimized difficulty level, tailored to the specific IPC regime. Moreover, through partial updates, it maintains the unique features of some real samples in the unchanged portion of the synthetic set. Meanwhile, the representative features of a whole dataset are translated through distillation to the rest of the images.
SelMatch sets a new benchmark in performance, outperforming traditional dataset distillation and sample selection methods across various subset ratios on the CIFAR-10/100 and Tiny ImageNet datasets.

Our SelMatch approach holds two primary limitations. First, it requires predefined difficulty scores. While our selection strategy (sliding window) avoids a computationally expensive optimization process, it depends on the existence of difficulty scores for training samples. If pre-computed difficulty scores are unavailable, additional computational overhead is incurred. In our study, we used measures like C-score \cite{cscore} and Forgetting score \cite{forgetting}, both of which demand significant computational resources. Alternatively, EL2N \cite{EL2N} can be employed to assess difficulty with minimal computation by training the network for only a few epochs and quantifying the difficulty of each sample using the initial loss. We provide the corresponding results in Appendix \ref{subsec:ablation_score}.

Second, SelMatch's hyperparameter tuning introduces additional computation overhead. SelMatch introduces two hyperparameters, the distillation portion ($\alpha$) and the window starting point ($\beta$), to regulate the incorporation of hard patterns in the distilled set. The optimal values of these hyperparameters vary with the dataset and IPC, necessitating tuning for each setting. We provide tuning guidance in Appendix \ref{sec:tuning_guidance} to help find optimal values more efficiently. 
Further developing a metric to estimate the optimal $\alpha$ and $\beta$ without the distillation process remains as an open challenge.

\section*{Impact Statement}

This paper introduces an effectively scalable dataset distillation method that extends beyond extremely small IPC scales. Our approach enables the training of neural networks in a manner that is both computationally and memory-efficient, with minimal to no performance loss compared to training with the full dataset. Significantly, our experimental results demonstrate that the synthetic dataset generated by our method generalizes well across various architectures. It proves effective not only in simpler ConvNet architectures, traditionally used for dataset distillation testing, but also in more complex models like ResNet. This versatility highlights the potential of our distilled dataset in maintaining high performance, even when applied to sophisticated neural network architectures. Such adaptability is crucial for advancing dataset distillation, ensuring efficient and robust training across diverse architectures.

\section*{Acknowledgements}

This research was supported by the National Research Foundation of Korea under grant 2021R1C1C11008539.


\bibliography{main_SelMatch}

\begin{thebibliography}{37}
\providecommand{\natexlab}[1]{#1}
\providecommand{\url}[1]{\texttt{#1}}
\expandafter\ifx\csname urlstyle\endcsname\relax
  \providecommand{\doi}[1]{doi: #1}\else
  \providecommand{\doi}{doi: \begingroup \urlstyle{rm}\Url}\fi

\bibitem[Cazenavette et~al.(2022)Cazenavette, Wang, Torralba, Efros, and
  Zhu]{tm}
Cazenavette, G., Wang, T., Torralba, A., Efros, A.~A., and Zhu, J.-Y.
\newblock Dataset distillation by matching training trajectories.
\newblock In \emph{Proceedings of the IEEE/CVF Conference on Computer Vision
  and Pattern Recognition}, 2022.

\bibitem[Chen et~al.(2010)Chen, Welling, and Smola]{herding}
Chen, Y., Welling, M., and Smola, A.
\newblock Super-samples from kernel herding.
\newblock In \emph{Proceedings of the Twenty-Sixth Conference on Uncertainty in
  Artificial Intelligence}, pp.\  109--116, 2010.

\bibitem[Cui et~al.(2022)Cui, Wang, Si, and Hsieh]{dcbench}
Cui, J., Wang, R., Si, S., and Hsieh, C.-J.
\newblock Dc-bench: Dataset condensation benchmark.
\newblock \emph{Advances in Neural Information Processing Systems}, 2022.

\bibitem[Cui et~al.(2023)Cui, Wang, Si, and Hsieh]{tesla}
Cui, J., Wang, R., Si, S., and Hsieh, C.-J.
\newblock Scaling up dataset distillation to imagenet-1k with constant memory.
\newblock In \emph{International Conference on Machine Learning}, 2023.

\bibitem[Deng et~al.(2009)Deng, Dong, Socher, Li, Li, and Fei-Fei]{imagenet}
Deng, J., Dong, W., Socher, R., Li, L.-J., Li, K., and Fei-Fei, L.
\newblock Imagenet: A large-scale hierarchical image database.
\newblock In \emph{2009 IEEE conference on computer vision and pattern
  recognition}, 2009.

\bibitem[Du et~al.(2023)Du, Jiang, Tan, Zhou, and Li]{ftd}
Du, J., Jiang, Y., Tan, V.~Y., Zhou, J.~T., and Li, H.
\newblock Minimizing the accumulated trajectory error to improve dataset
  distillation.
\newblock In \emph{Proceedings of the IEEE/CVF Conference on Computer Vision
  and Pattern Recognition}, 2023.

\bibitem[Du et~al.(2024)Du, Shi, and Zhou]{seqmatch}
Du, J., Shi, Q., and Zhou, J.~T.
\newblock Sequential subset matching for dataset distillation.
\newblock \emph{Advances in Neural Information Processing Systems}, 2024.

\bibitem[Gidaris \& Komodakis(2018)Gidaris and Komodakis]{convnet}
Gidaris, S. and Komodakis, N.
\newblock Dynamic few-shot visual learning without forgetting.
\newblock In \emph{Proceedings of the IEEE conference on computer vision and
  pattern recognition}, 2018.

\bibitem[Guo et~al.(2023)Guo, Wang, Cazenavette, LI, Zhang, and You]{datm}
Guo, Z., Wang, K., Cazenavette, G., LI, H., Zhang, K., and You, Y.
\newblock Towards lossless dataset distillation via difficulty-aligned
  trajectory matching.
\newblock In \emph{The Twelfth International Conference on Learning
  Representations}, 2023.

\bibitem[He et~al.(2016)He, Zhang, Ren, and Sun]{resnet}
He, K., Zhang, X., Ren, S., and Sun, J.
\newblock Deep residual learning for image recognition.
\newblock In \emph{IEEE Conference on Computer Vision and Pattern Recognition},
  2016.

\bibitem[Ioffe \& Szegedy(2015)Ioffe and Szegedy]{batchnorm}
Ioffe, S. and Szegedy, C.
\newblock Batch normalization: Accelerating deep network training by reducing
  internal covariate shift.
\newblock In \emph{International conference on machine learning}, 2015.

\bibitem[Jiang et~al.(2021)Jiang, Zhang, Talwar, and Mozer]{cscore}
Jiang, Z., Zhang, C., Talwar, K., and Mozer, M.~C.
\newblock Characterizing structural regularities of labeled data in
  overparameterized models.
\newblock In \emph{International Conference on Machine Learning}, 2021.

\bibitem[Killamsetty et~al.(2021{\natexlab{a}})Killamsetty, Durga,
  Ramakrishnan, De, and Iyer]{gradmatch}
Killamsetty, K., Durga, S., Ramakrishnan, G., De, A., and Iyer, R.
\newblock Grad-match: Gradient matching based data subset selection for
  efficient deep model training.
\newblock In \emph{International Conference on Machine Learning},
  2021{\natexlab{a}}.

\bibitem[Killamsetty et~al.(2021{\natexlab{b}})Killamsetty, Sivasubramanian,
  Ramakrishnan, and Iyer]{glister}
Killamsetty, K., Sivasubramanian, D., Ramakrishnan, G., and Iyer, R.
\newblock Glister: Generalization based data subset selection for efficient and
  robust learning.
\newblock In \emph{Association for the Advancement of Artificial Intelligence},
  2021{\natexlab{b}}.

\bibitem[Koh \& Liang(2017)Koh and Liang]{infl_func}
Koh, P.~W. and Liang, P.
\newblock Understanding black-box predictions via influence functions.
\newblock In \emph{Advances in Neural Information Processing Systems}, 2017.

\bibitem[Kolossov et~al.(2023)Kolossov, Montanari, and
  Tandon]{kolossov2023towards}
Kolossov, G., Montanari, A., and Tandon, P.
\newblock Towards a statistical theory of data selection under weak
  supervision.
\newblock In \emph{The Twelfth International Conference on Learning
  Representations}, 2023.

\bibitem[Krizhevsky et~al.(2012)Krizhevsky, Sutskever, and
  Hinton]{krizhevsky2012imagenet}
Krizhevsky, A., Sutskever, I., and Hinton, G.~E.
\newblock Imagenet classification with deep convolutional neural networks.
\newblock \emph{Advances in neural information processing systems}, 2012.

\bibitem[Krizhevsky et~al.(2009)]{cifar}
Krizhevsky, A. et~al.
\newblock Learning multiple layers of features from tiny images.
\newblock 2009.

\bibitem[Le \& Yang(2015)Le and Yang]{tiny}
Le, Y. and Yang, X.
\newblock Tiny imagenet visual recognition challenge.
\newblock \emph{CS 231N}, 7\penalty0 (7):\penalty0 3, 2015.

\bibitem[Mirzasoleiman et~al.(2020)Mirzasoleiman, Bilmes, and Leskovec]{craig}
Mirzasoleiman, B., Bilmes, J., and Leskovec, J.
\newblock Coresets for data-efficient training of machine learning models.
\newblock In \emph{International Conference on Machine Learning}, 2020.

\bibitem[Nguyen et~al.(2020)Nguyen, Chen, and Lee]{kip}
Nguyen, T., Chen, Z., and Lee, J.
\newblock Dataset meta-learning from kernel ridge-regression.
\newblock In \emph{International Conference on Learning Representations}, 2020.

\bibitem[Nguyen et~al.(2021)Nguyen, Novak, Xiao, and Lee]{kip2}
Nguyen, T., Novak, R., Xiao, L., and Lee, J.
\newblock Dataset distillation with infinitely wide convolutional networks.
\newblock \emph{Advances in Neural Information Processing Systems}, 2021.

\bibitem[Paul et~al.(2021)Paul, Ganguli, and Dziugaite]{EL2N}
Paul, M., Ganguli, S., and Dziugaite, G.~K.
\newblock Deep learning on a data diet: Finding important examples early in
  training.
\newblock In \emph{Advances in Neural Information Processing Systems}, 2021.

\bibitem[Pruthi et~al.(2020)Pruthi, Liu, Kale, and Sundararajan]{TracIn}
Pruthi, G., Liu, F., Kale, S., and Sundararajan, M.
\newblock Estimating training data influence by tracing gradient descent.
\newblock In \emph{Advances in Neural Information Processing Systems}, 2020.

\bibitem[Sener \& Savarese(2018)Sener and Savarese]{kcenter}
Sener, O. and Savarese, S.
\newblock Active learning for convolutional neural networks: A core-set
  approach.
\newblock In \emph{International Conference on Learning Representations}, 2018.

\bibitem[Simonyan \& Zisserman(2014)Simonyan and Zisserman]{vgg}
Simonyan, K. and Zisserman, A.
\newblock Very deep convolutional networks for large-scale image recognition.
\newblock \emph{arXiv preprint arXiv:1409.1556}, 2014.

\bibitem[Sorscher et~al.(2022)Sorscher, Geirhos, Shekhar, Ganguli, and
  Morcos]{sorscher2022beyond}
Sorscher, B., Geirhos, R., Shekhar, S., Ganguli, S., and Morcos, A.
\newblock Beyond neural scaling laws: beating power law scaling via data
  pruning.
\newblock \emph{Advances in Neural Information Processing Systems}, 2022.

\bibitem[Toneva et~al.(2018)Toneva, Sordoni, Combes, Trischler, Bengio, and
  Gordon]{forgetting}
Toneva, M., Sordoni, A., Combes, R. T.~d., Trischler, A., Bengio, Y., and
  Gordon, G.~J.
\newblock An empirical study of example forgetting during deep neural network
  learning.
\newblock \emph{arXiv preprint arXiv:1812.05159}, 2018.

\bibitem[Ulyanov et~al.(2016)Ulyanov, Vedaldi, and Lempitsky]{instancenorm}
Ulyanov, D., Vedaldi, A., and Lempitsky, V.
\newblock Instance normalization: The missing ingredient for fast stylization.
\newblock \emph{arXiv preprint arXiv:1607.08022}, 2016.

\bibitem[Van~der Maaten \& Hinton(2008)Van~der Maaten and Hinton]{tsne}
Van~der Maaten, L. and Hinton, G.
\newblock Visualizing data using t-sne.
\newblock \emph{Journal of machine learning research}, 2008.

\bibitem[Wang et~al.(2018)Wang, Zhu, Torralba, and Efros]{dd}
Wang, T., Zhu, J.-Y., Torralba, A., and Efros, A.~A.
\newblock Dataset distillation.
\newblock \emph{arXiv preprint arXiv:1811.10959}, 2018.

\bibitem[Yin et~al.(2024)Yin, Xing, and Shen]{sre2l}
Yin, Z., Xing, E., and Shen, Z.
\newblock Squeeze, recover and relabel: Dataset condensation at imagenet scale
  from a new perspective.
\newblock \emph{Advances in Neural Information Processing Systems}, 2024.

\bibitem[Zhao \& Bilen(2021)Zhao and Bilen]{dsa}
Zhao, B. and Bilen, H.
\newblock Dataset condensation with differentiable siamese augmentation.
\newblock In \emph{International Conference on Machine Learning}, 2021.

\bibitem[Zhao \& Bilen(2023)Zhao and Bilen]{dm}
Zhao, B. and Bilen, H.
\newblock Dataset condensation with distribution matching.
\newblock In \emph{Proceedings of the IEEE/CVF Winter Conference on
  Applications of Computer Vision (WACV)}, 2023.

\bibitem[Zhao et~al.(2020)Zhao, Mopuri, and Bilen]{dc}
Zhao, B., Mopuri, K.~R., and Bilen, H.
\newblock Dataset condensation with gradient matching.
\newblock In \emph{International Conference on Learning Representations}, 2020.

\bibitem[Zhou et~al.(2023)Zhou, Wang, Gu, Peng, Lian, Zhang, You, and Feng]{dq}
Zhou, D., Wang, K., Gu, J., Peng, X., Lian, D., Zhang, Y., You, Y., and Feng,
  J.
\newblock Dataset quantization.
\newblock In \emph{Proceedings of the IEEE/CVF International Conference on
  Computer Vision}, 2023.

\bibitem[Zhou et~al.(2022)Zhou, Nezhadarya, and Ba]{frepo}
Zhou, Y., Nezhadarya, E., and Ba, J.
\newblock Dataset distillation using neural feature regression.
\newblock \emph{Advances in Neural Information Processing Systems}, 2022.

\end{thebibliography}
\bibliographystyle{icml2024}

\newpage
\appendix
\onecolumn
\section{Full Algorithm: SelMatch}
\begin{algorithm}[htb!]
\caption{Dataset Distillation via Selection and Matching (SelMatch)}
\label{alg:main}
\begin{algorithmic}
\STATE {\bfseries Input:} Full training set $\mathcal{D}_\textrm{real}$, difficulty score $\{s_i\}_{i=1}^{|\mathcal{D}_\textrm{real}|}$, number of classes $C$, images per class $\mathrm{IPC}$, set of expert trajectories $\{ \tau^*_i \}$, Maximum start epoch $T^+$, Differentiable augmentation function $\mathcal{A}$, number of updates for student network  $N$, number of updates from the chosen start epoch for target expert trajectory $M$, distillation portion $\alpha\in[0,1]$, window starting point $\beta\in[0,1]$.
\STATE 
\STATE $\triangleright$ Selection-based initialization with sliding-window algorithm.
\STATE Sort $\mathcal{D}_\textrm{real} = \{(x_i, y_i)\}_{i=1}^{|\mathcal{D}_\textrm{real}|}$ by difficulty score $\{s_i\}_{i=1}^{|\mathcal{D}_\textrm{real}|}$ in descending order.
\STATE Re-order it while guaranteeing $y_i=(i\mod C)$ and $s_i\geq s_{i+kC}$ for $k\in \mathbb{Z}^+$.
\STATE $m \gets \lceil \beta \times |\mathcal{D}_\textrm{real}| \rceil$ \hfill $\triangleright$ prune the hardest $\beta \times 100$\% samples
    \STATE $\mathcal{D}_\textrm{select} \gets  \{(x_i, y_i)\}_{i=m}^{m + \lceil (1-\alpha) \times \mathrm{IPC} \times C \rceil}$
    \STATE $\mathcal{D}_\textrm{distill} \gets  \{(x_i, y_i)\}_{i=m + \lceil (1-\alpha) \times \mathrm{IPC}\times C \rceil+1}^{m + \mathrm{IPC}\times C}$
\STATE 
\STATE $\triangleright$ Matching with partial update.
\STATE Freeze $\mathcal{D}_\textrm{select}$.
\REPEAT
\STATE Sample expert trajectory $\tau^* \sim \{ \tau^*_i \}$ with $\tau^* = \{ \theta^*_t \}_0^T$
\STATE Sample start epoch $t$, where $t \leq T^+$
\STATE $\hat{\theta}_t \gets \theta_t^*$ \hfill $\triangleright$ Initialize student network
\FOR{$i=1$ {\bfseries to} $N$}
    \STATE $b_{t+i} \sim (\mathcal{D}_\textrm{select} \cup \mathcal{D}_\textrm{distill})$ \hfill $\triangleright$ Sample a mini-batch from the entire synthetic set
    \STATE $\hat{\theta}_{t+i} \gets \hat{\theta}_{t+i-1} - \eta \nabla l(\mathcal{A}(b_{t+i});\hat{\theta}_{t+i-1})$ \hfill $\triangleright$ Update student network w.r.t. classification loss
\ENDFOR
\STATE Compute matching loss $\mathcal{L}$ between $\hat{\theta}_{t+N}$ and $\theta^*_{t+M}$ with \eqref{matching loss}.
\STATE Update $\mathcal{D}_\textrm{distill}$ and $\eta$ with respect to $\mathcal{L}$.
\UNTIL Converge

\STATE {\bfseries Output:} Synthetic dataset $\mathcal{D}_\textrm{syn} = \mathcal{D}_\textrm{select} \cup \mathcal{D}_\textrm{distill}$
\end{algorithmic}
\end{algorithm}

\section{Implementation Details}
\label{sec:implement_detail}

\subsection{SelMatch and Reproduction of MTT}
In our experiments, we follow the hyperparameters outlined in the original MTT paper \cite{tm} for generating expert trajectories. However, we introduce some modifications to the hyperparameters related to the distillation process. As discussed in Section \ref{subsec:ablation_study}, we employ the ConvNet architecture with batch normalization (ConvNet-BN) for distillation and extend the maximum start epoch $T^+$ to 80, as this adjustment has been shown to enhance performance in scenarios with large IPC. We also reproduce MTT using ConvNet-BN and the increased $T^+$. Additionally, on CIFAR-100 with IPC=50, MTT updates student networks using full synthetic set instead of sampling a mini-batch and uses a large number of student updates $N=80$. Due to the linear scaling of required memory with respect to $N$ and synthetic batch size $|b|$, distillation with these settings becomes exceedingly resource-intensive. To address this, we impose constraints on these hyperparameters, setting $N=55$ on CIFAR-10/100 (20 on Tiny ImageNet) and $|b|=125$ across all settings in both SelMatch and the reproduction of MTT, for practicality and efficiency. Moreover, MTT applies ZCA whitening in certain cases but omits its use in others. Given that the ablation results in the MTT paper suggest that ZCA whitening has a negligible effect on performance when IPC is not very small, we perform distillation without employing ZCA in all settings including reproduction of MTT, for the sake of simplicity. 

For SelMatch, in sorting samples based on their difficulty scores, we utilize pre-computed C-score \cite{cscore} on CIFAR-10/100 and use Forgetting score \cite{forgetting} on Tiny Imagenet as the difficulty score. Then, we find an optimal window starting point $\beta$ for each dataset and subset ratio by the sliding window algorithm. After finding the optimal $\beta$, we tune the values of distillation portion $\alpha$ with the fixed $\beta$.  It is worth noting that we can tune $\alpha$ and $\beta$ more efficiently than through grid search based on our insights on the optimal $\alpha$ and $\beta$. More details are provided in Appendix \ref{sec:tuning_guidance}.  We report the hyperparameters used for our main result in Table \ref{appendix:hyperparams}. For both SelMatch and MTT, we distill for 10,000 iterations to ensure convergence. All other hyperparameters are remained unchanged from the original MTT paper.

\begin{table*}[t]
\caption{Hyper-parameters used for SelMatch.}
\label{appendix:hyperparams}
\vskip 0.15in
\begin{center}
\begin{scriptsize}
\begin{tabular}{cc|cccccc}
\toprule
\multirow{2}{*}{Dataset} & \multirow{2}{*}{IPC} & Student Updates & Expert Epochs & Synthetic Batchsize & Distillation Portion & Window Start & Learning Rate \\
& & ($N$) & ($M$) & ($|b|$) & ($\alpha$) & ($\beta \%$) & (pixel) \\
\midrule
\multirow{4}{*}{CIFAR-10} & 250 & \multirow{4}{*}{55} & \multirow{4}{*}{2} & \multirow{4}{*}{125} & 0.6 & 50 & 1000 \\
& 500 & & & & 0.2 & 30 & 1000 \\
& 1000 & & & & 0.1 & 20 & 10000 \\
& 1500 & & & & 0.1 & 20 & 10000 \\
\midrule
\multirow{4}{*}{CIFAR-100} & 25 & \multirow{4}{*}{55} & \multirow{4}{*}{2} & \multirow{4}{*}{125} & 0.8 & 80 & 1000 \\
& 50 & & & & 0.6 & 70 & 1000 \\
& 100 & & & & 0.3 & 60 & 1000 \\
& 150 & & & & 0.2 & 40 & 1000 \\
\midrule
\multirow{2}{*}{Tiny} & 50 & \multirow{2}{*}{20} & \multirow{2}{*}{2} & \multirow{2}{*}{125} & 0.6 & 80 & 1000 \\
& 100 & & & & 0.5 & 70 & 1000 \\
\bottomrule
\end{tabular}
\end{scriptsize}
\end{center}
\vskip -0.1in
\end{table*}

\subsection{Reproduction of Other Baselines} 

\paragraph{DSA / DM} We reproduce DSA \cite{dsa} / DM \cite{dm} baselines only on CIFAR-10 with IPC ranging from 250 to 1000 and CIFAR-100 with IPC ranging from 25 to 50, due to scalability issue. For Tiny ImageNet with IPC=50, we use distilled dataset provided by original papers without reproduction. As hyper-parameters for large IPC settings are not reported in the original papers, we set the number of inner optimizations and outer optimizations both to 10 following DC-BENCH \cite{dcbench}. We distill the dataset for 1,000 iterations with learning rate 0.1 for DSA and 10 for DM. Following the original papers, no ZCA whitening is applied for DSA and DM.

\paragraph{FTD / DATM} Since distilled datasets for FTD \cite{ftd} are not provided, we reproduce FTD baselines in all settings. As hyper-parameters for large IPC scales are not given, we set hyper-parameters to the same values used in the original paper for the corresponding dataset of the largest IPC setting  (IPC=50 for CIFAR-10/100 and IPC=10 on Tiny ImageNet). Consistent to SelMatch and MTT, we restricte $N$ to 55 on CIFAR-10/100 (20 on Tiny ImageNet) and $|b|$ to 125. We perform distillation on 5,000 iterations with ZCA whitening and exponential moving average (EMA), following the original paper. For DATM \cite{datm}, we evaluate with distilled datasets provided by original paper. Note that these datasets are distilled with much larger memory and computation cost ($N=80$, $|b|=250$ or 1000) and ZCA whitening.

\subsection{Evaluation of Distilled Dataset} 
We evaluate all methods by training ResNet18 network with batch normalization using SGD optimizer with momentum 0.9, weight decay 5e-4, learning rate 0.1, and batch size 128. For SelMatch and MTT, we utilize cosine annealing scheduler. For other baselines, we employ step scheduler following the original papers. Additionally, we apply random cropping padded with 4 pixels and random horizontal flipping on $\mathcal{D}_\mathrm{select}$ for combined augmentation of our method, which is typically used for subset selection method.

\section{More Experimental Results}

\subsection{Sliding Window Experiment}

\begin{figure*}[htb!]
\centering
    \subfigure[CIFAR-10]
    {\includegraphics[width=0.32\linewidth]{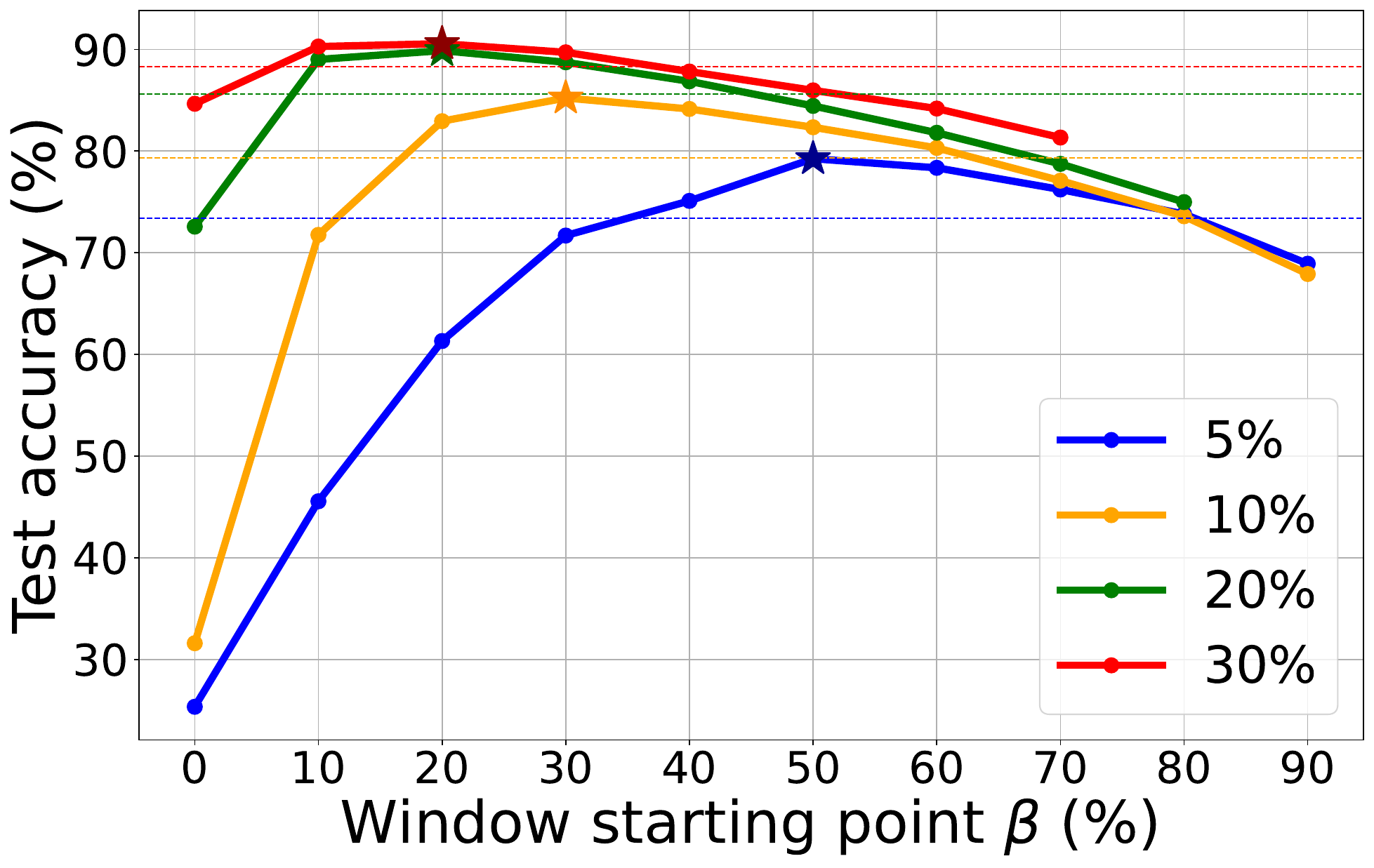}}
    \subfigure[CIFAR-100]
    {\includegraphics[width=0.32\linewidth]{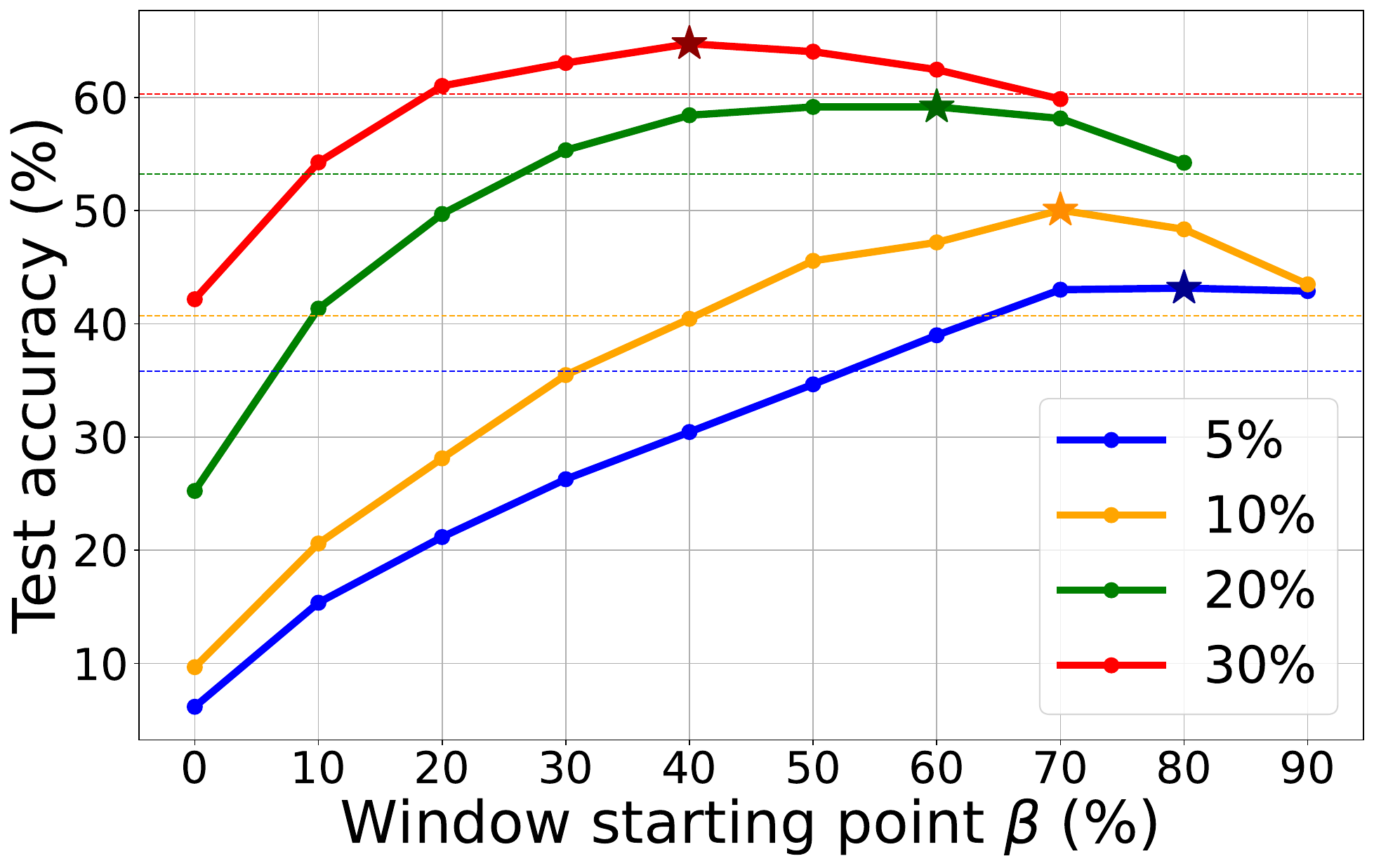}}
    \subfigure[Tiny ImageNet]
    {\includegraphics[width=0.32\linewidth]{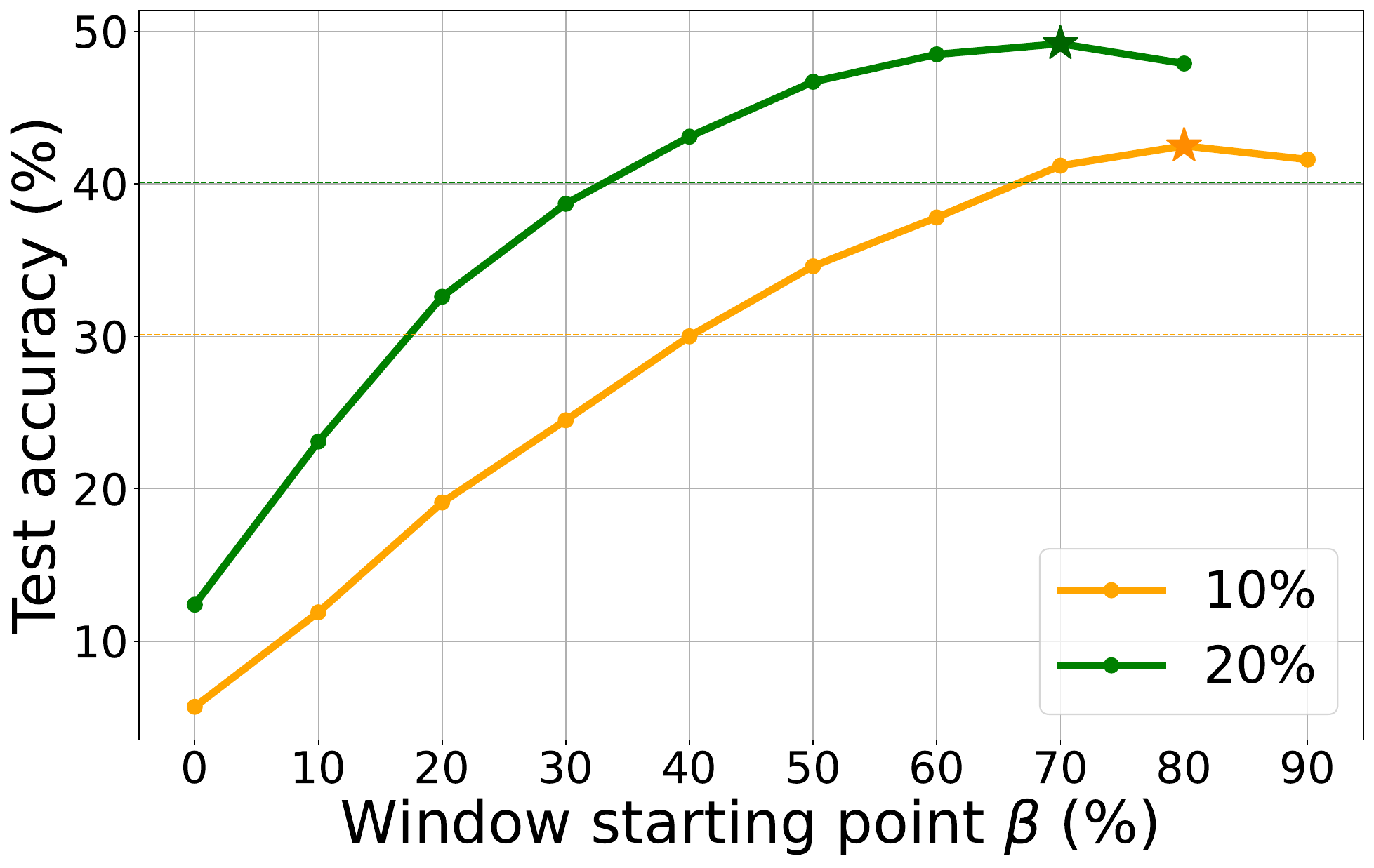}}
    \caption{The result of sliding window experiment on CIFAR-10/100 and Tiny ImageNet.}
    \label{fig:appendix_window}
\end{figure*}

To validate the consistency of our sliding-window algorithm (described in Sec. \ref{sec:method}) across diverse datasets, we show additional results on CIFAR-100 and Tiny ImageNet, as illustrated in Figure \ref{fig:appendix_window}. As the complexity of dataset increases (from CIFAR to Tiny ImageNet), the optimal $\beta$ achieving the highest test accuracy tends to increase, implying that the optimal window subset for selection-based initialization includes easy samples for complex dataset. As the subset ratio increases, on the other hand, the optimal $\beta$ decreases, indicating the importance of incorporating hard and complex samples in the increased IPC. 

\subsection{Coverage Analysis}
\begin{figure*}[htb!]
  \begin{center}
      \includegraphics[width=.95\linewidth]{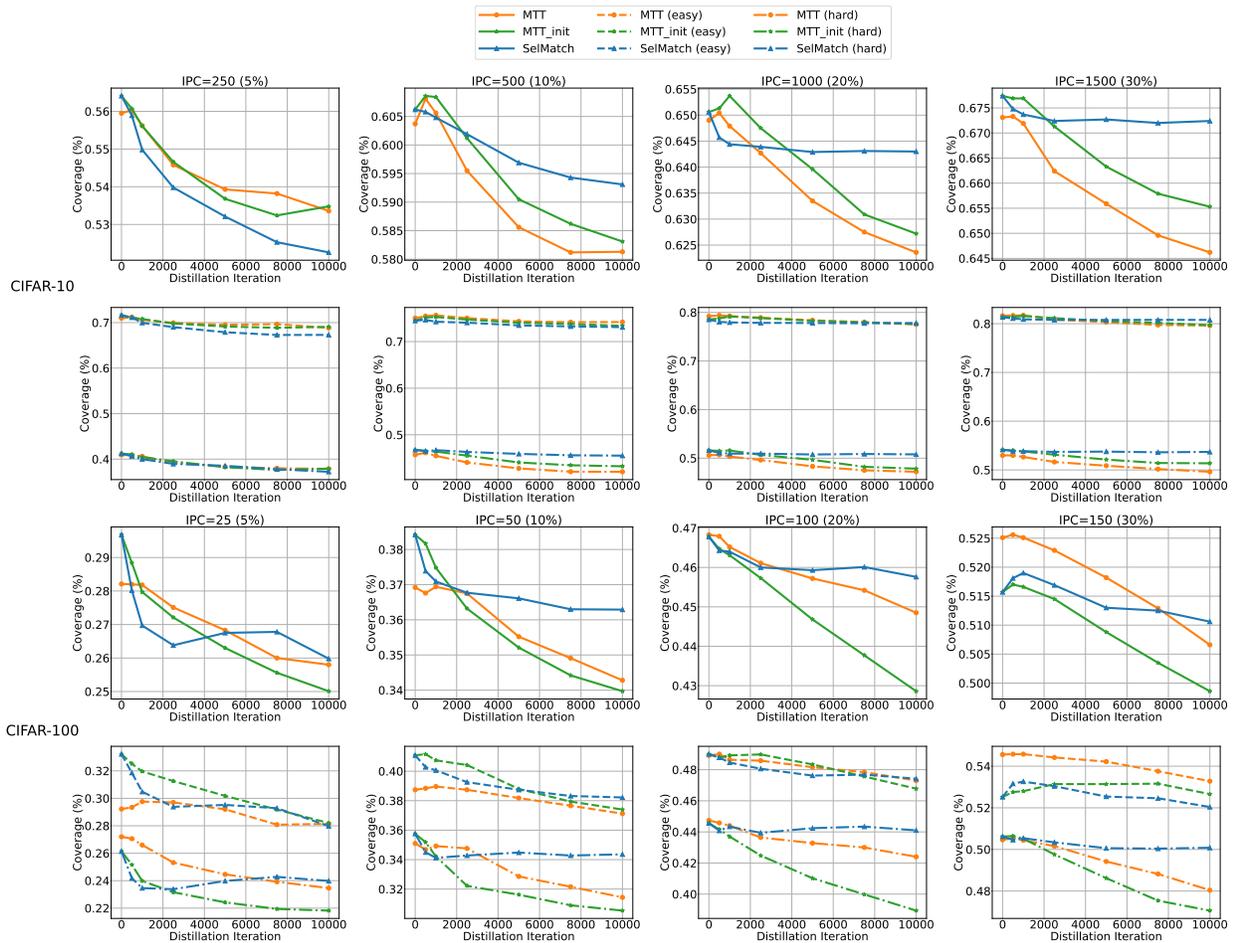}
  \end{center}
  \caption{Change of coverage throughout the distillation process. The first two rows and the last two rows show the results for CIFAR-10 and CIFAR-100, respectively, across various IPC settings. In each pair of rows, the first row plots coverage on the entire dataset and the second row plots coverage on easy vs. hard groups.}
  \label{fig:appendix_iter_coverage}
\end{figure*}

We present more results on coverage analysis, similar to that presented in Sec. \ref{sec:coverage}.  `MTT' represents the original algorithm, initialized with random samples, while `MTT\_init' refers to MTT but initialized with the optimal window subset $\mathcal{D}_\mathrm{initial}$, as like SelMatch.  As can be observed in Figure \ref{fig:appendix_iter_coverage}, both for MTT and MTT\_init, updating all samples during distillation results in rapid diminish of coverage as the distillation iteration increases, especially on large IPC. This coverage diminish is caused because traditional MTT overly changes rare and unique features of real samples into easy and representative patterns. On the other hand, the coverage by SelMatch remains stable due to the partial updates of SelMatch, which maintains the unique and rare features of selected hard samples within the synthetic dataset. 

\section{Further Ablation Studies}

\subsection{Other Difficulty Scores}
\label{subsec:ablation_score}

\begin{figure*}[htb!]
\centering
    \subfigure[IPC=50]
    {\includegraphics[width=0.32\linewidth]{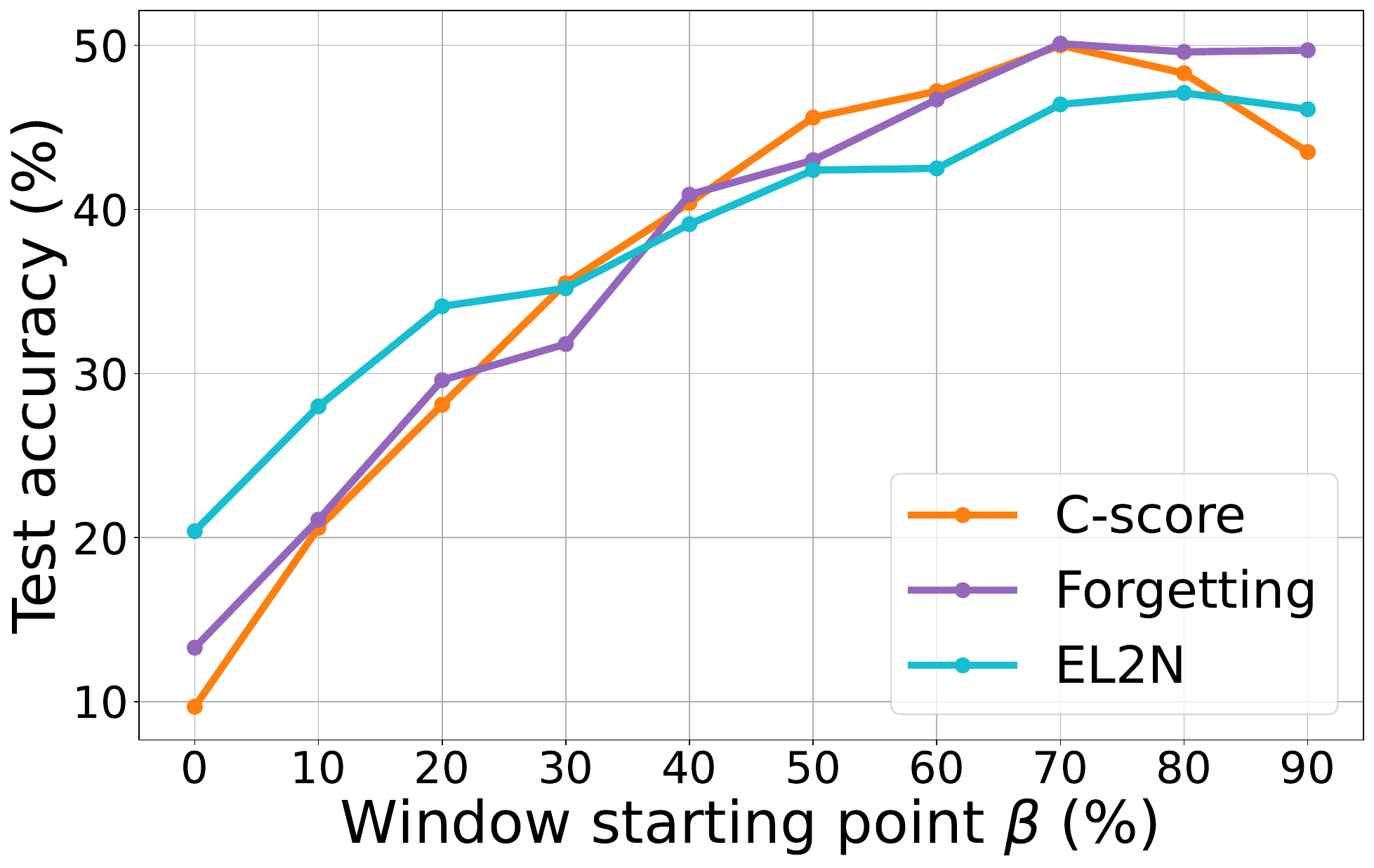}}
    \quad
    \subfigure[IPC=100]
    {\includegraphics[width=0.33\linewidth]{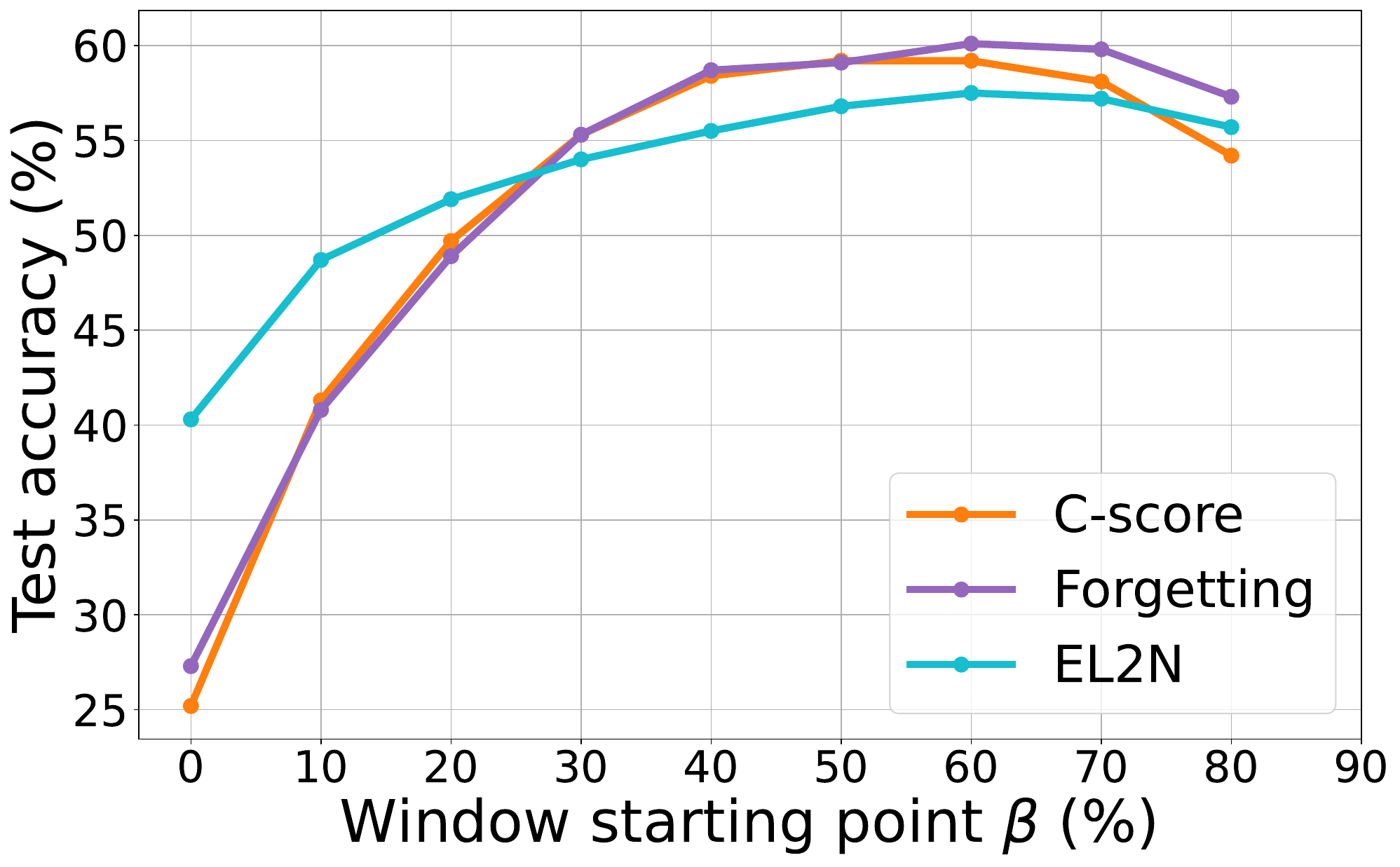}}
    \caption{The result of sliding window experiment on CIFAR-100 with different difficulty scores.}
    \label{fig:appendix_window_other_scores}
\end{figure*}

SelMatch requires a measure of the difficulty of training samples to determine the appropriate difficulty level. For this purpose, we leverage C-score for CIFAR-10/100 and Forgetting score metrics, both of which necessitate substantial computational resources for computation, particularly the C-score. In cases where pre-computed difficulty scores were unavailable, we explored the use of an alternative difficulty metric known as EL2N \cite{EL2N}. EL2N quantifies the initial loss during training and consequently necessitates only a few epochs of network training for computation. We conducted sliding window experiments with different scores on CIFAR-100 with IPC=50, 100 and present the results in Figure \ref{fig:appendix_window_other_scores}. All three test accuracy - difficulty curves using different difficulty scores exhibit a similar shape, but the curve with the EL2N score appears more flat, resulting in a lower best test accuracy. We conjecture that EL2N has less power to measure the samples' difficulty because it utilizes only information from the early training phase. In Table \ref{tab:appendix_ablation_score}, we present the performance of SelMatch when utilizing these three different difficulty scores. We observe that using the Forgetting score also leads to high performance comparable to using the C-score, which is leveraged for the main result (Table \ref{tab:main_result}). This verifies the robustness of our method to the choice of difficulty metric. However, when using EL2N score, there is slight performance degradation, which results from the lower performance of the initialization point. This result can be viewed as a trade-off between computation cost and performance.

\begin{table}[htb!]
\caption{Performance of SelMatch using other different sample difficulty measures on CIFAR-100 with IPC=50 and 100.}
\label{tab:appendix_ablation_score}
\vskip 0.15in
\begin{center}
\begin{small}
\begin{tabular}{c|cccc}
\toprule
IPC & C-score & Forgetting & EL2N \\
\midrule
50 & 54.5 & 54.2 & 52.9 \\
100 & 62.4 & 62.9 & 61.6 \\
\bottomrule
\end{tabular}
\end{small}
\end{center}
\vskip -0.1in
\end{table}

\subsection{Other Distillation Methods}
\label{subsec:other_baselines}

For implementation, we utilize MTT \cite{tm} as the base distillation method. However, our selection-based initialization and partial update strategies are independent of the base dataset distillation method and can be applied to other dataset distillation methods. To explore SelMatch’s adaptability to different baseline methods, we implemented SelMatch on two other baselines: DSA\cite{dsa} and SRe2L\cite{sre2l}, and evaluated performance on CIFAR-100 dataset. DSA updates the synthetic set to make its gradient similar to the gradient of the real dataset. SRe2L decouples the bi-level optimization of the model and synthetic set, allowing for the use of larger models on large-scale datasets. We report the results on DSA and SRe2L in Table \ref{appendix:ablation_dsa} and Table \ref{appendix:ablation_sre2l}, respectively. For SRe2L, we also include the performance of initializing with random real samples, as the original baseline initializes synthetic samples with Gaussian noise, for a fair comparison. The results demonstrate that both our initialization strategy and partial update mechanism significantly contribute to performance enhancement, underscoring the effectiveness of SelMatch in terms of its versatility and applicability.

\begin{table}[htb!]
\caption{Evaluation of SelMatch implemented on DSA baseline. The experiment is conducted on CIFAR-100 with IPC=25, 50. }
\label{appendix:ablation_dsa}
\vskip 0.15in
\begin{center}
\begin{small}
\begin{tabular}{c|c|cc}
\toprule
IPC & Baseline (DSA) & Select Init & Select Init + Partial Update (SelMatch) \\
\midrule
25 & 38.3 & 43.4 & 45.0 \\
50 & 43.6 & 49.8 & 50.8 \\
\bottomrule
\end{tabular}
\end{small}
\end{center}
\vskip -0.1in
\end{table}

\begin{table}[htb!]
\caption{Evaluation of SelMatch implemented on SRe2L baseline. The experiment is conducted on CIFAR-100 with IPC=50, 100.}
\label{appendix:ablation_sre2l}
\vskip 0.15in
\begin{center}
\begin{small}
\begin{tabular}{c|cc|cc}
\toprule
\multirow{2}{*}{IPC} & \multicolumn{2}{c|}{Baseline} & \multirow{2}{*}{Select Init} & Select Init + Partial Update \\ 
& Noise Init & Real Init & & (SelMatch) \\
\midrule
50 & 49.9 & 57.5 & 59.7 & 63.1 \\
100 & 57.1 & 61.4 & 64.2 & 67.1 \\
\bottomrule
\end{tabular}
\end{small}
\end{center}
\vskip -0.1in
\end{table}

\section{Tuning Guidance}
\label{sec:tuning_guidance}

\begin{figure*}[htb!]
\centering
    \subfigure[CIFAR-10]
    {\includegraphics[width=0.32\linewidth]{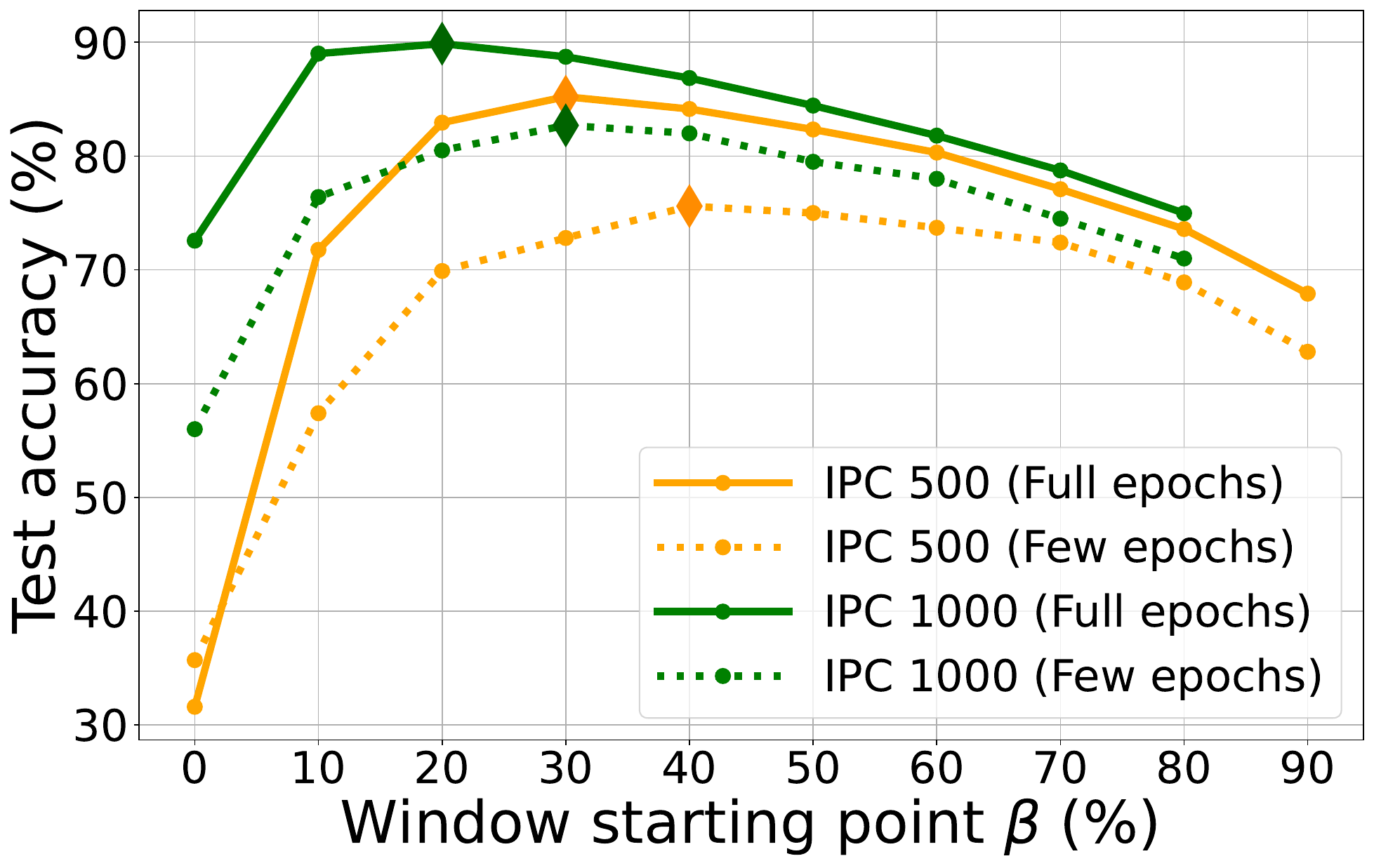}}
    \subfigure[CIFAR-100]
    {\includegraphics[width=0.32\linewidth]{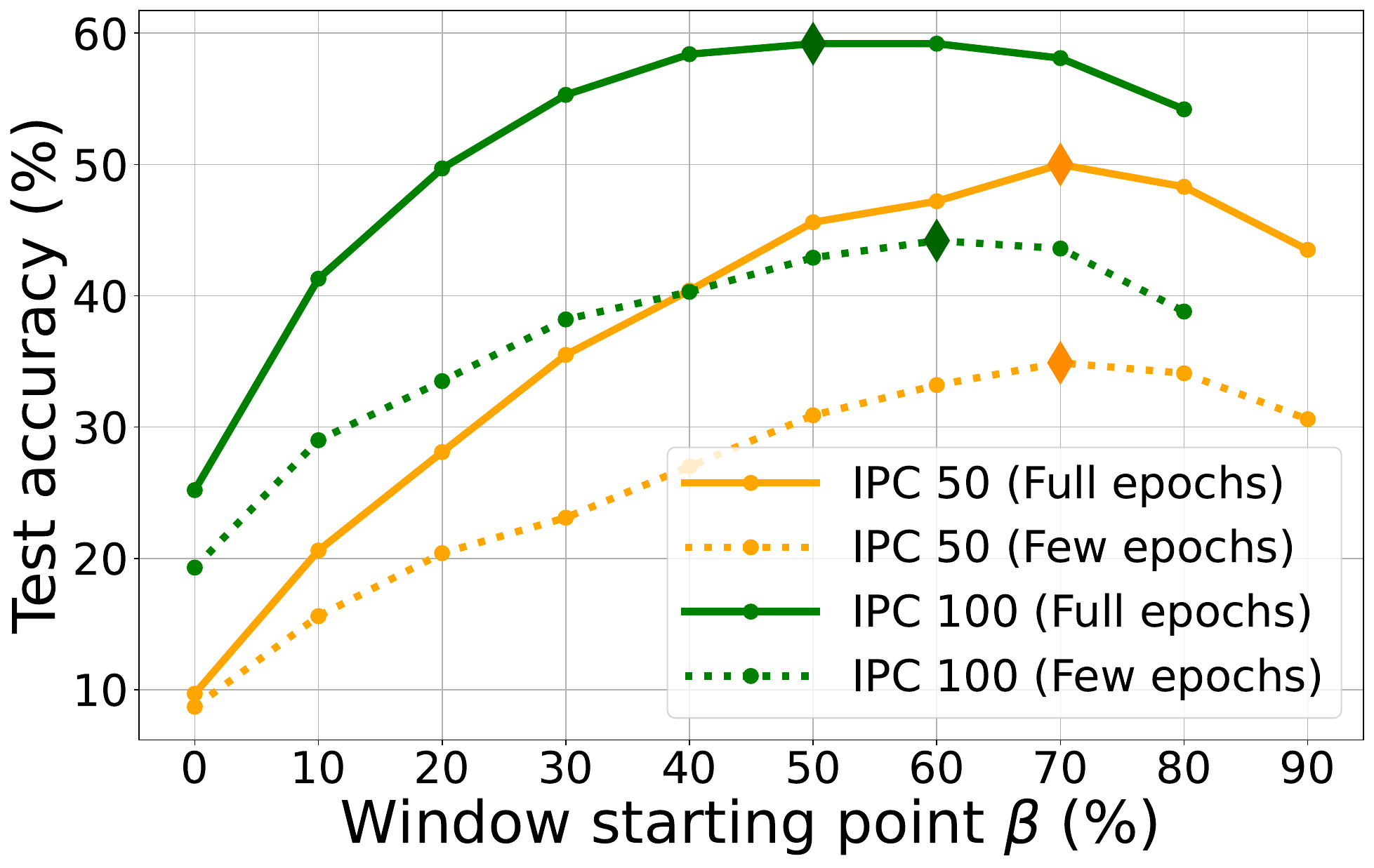}}
    \subfigure[Tiny ImageNet]
    {\includegraphics[width=0.32\linewidth]{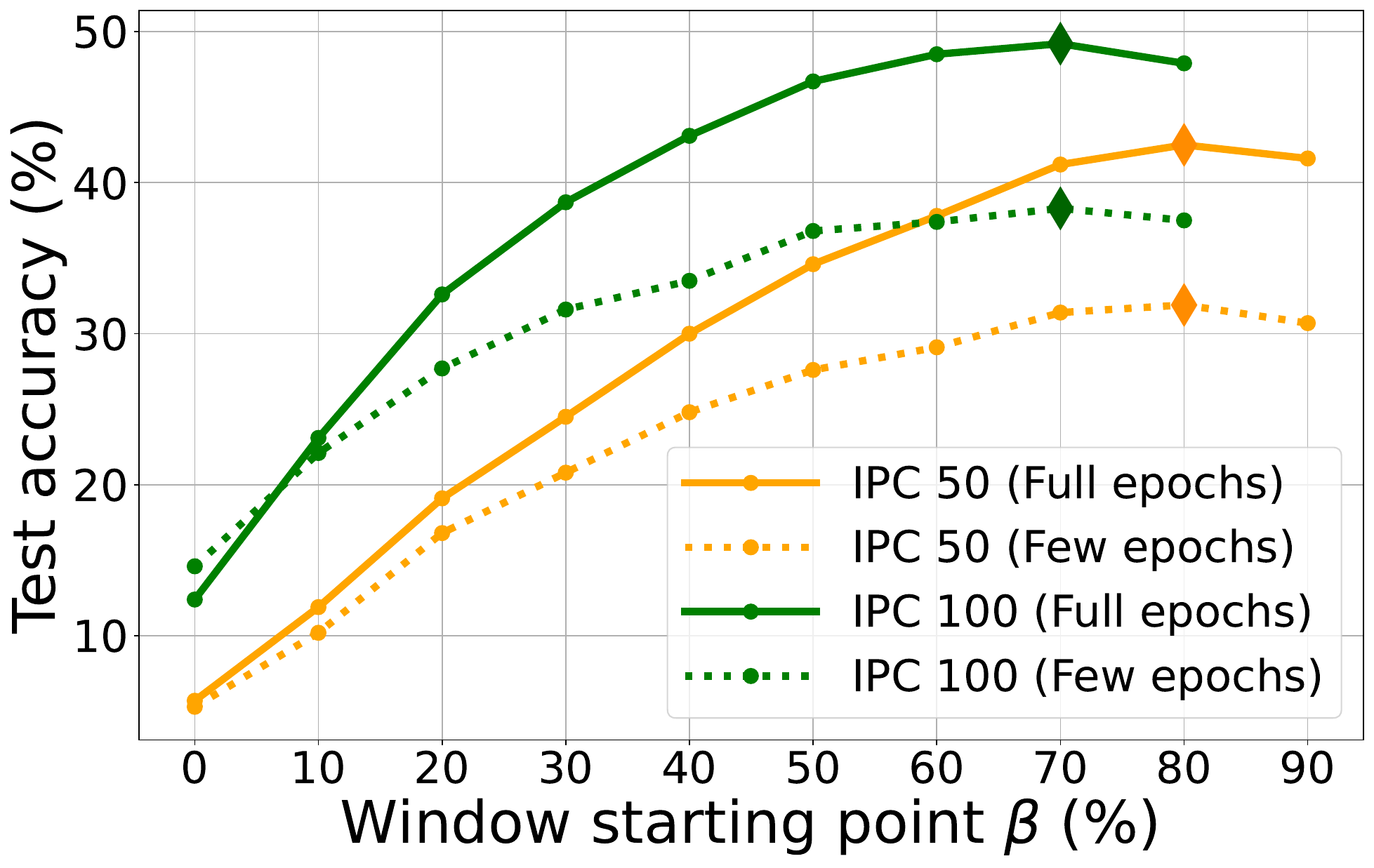}}
    \caption{Results of the sliding window experiment with two different epoch settings. In the first setting, the network is fully trained on each window subset (Full epochs), while in the second setting, the network is trained for only a few epochs (Few epochs).}
    \label{fig:appendix_few_epochs}
\end{figure*}

The implementation of SelMatch involves two critical hyperparameters: the distillation portion ($\alpha$) and the window starting point ($\beta$). However, determining the optimal values for $\alpha$ and $\beta$ incurs substantial computational overhead. To address this, we propose a more efficient approach for identifying these values.

To find the optimal window starting point ($\beta$), we suggest training the network on each window subset for only a few epochs rather than fully training the networks. To evaluate the efficacy of this approach, we compared the results of the sliding window algorithm using full epochs versus a reduced number of epochs, as shown in Figure \ref{fig:appendix_few_epochs}. Specifically, for the ``few epochs" setting, we used 20\% of the full number of epochs for CIFAR-10 and 10\% for CIFAR-100 and Tiny Imagenet. The figure demonstrates that the results from the sliding window algorithm with these two different epoch settings exhibit a high rank correlation and nearly identical optimal window.

Additionally, we can significantly reduce the search space by leveraging two important observations from Figures \ref{fig:window_selection} and \ref{fig:ablation_alpha}. First, the $\alpha$-test accuracy curve and the $\beta$-test accuracy curve exhibit a concave shape. Using this observation, we can eliminate more than half of the entire search space. For example, start searching at $\alpha$=0.5 and determine whether to increase or decrease the $\alpha$ value based on evaluations at adjacent points (e.g., $\alpha$ = 0.4 or 0.6). Once the direction for adjustment is established, continue searching until the test accuracy begins to decrease. Second, the optimal values for $\alpha$ and $\beta$ decrease as IPC increases. Using this observation, we can further reduce the search space. For instance, if we have already identified the optimal $\alpha$=0.3 on CIFAR-100 with IPC=100, then for larger IPC, we only need to search for $\alpha$ values smaller than 0.3.

\section{Visualization}

\begin{figure*}[htb!]
  \begin{center}
      \includegraphics[width=.95\linewidth]{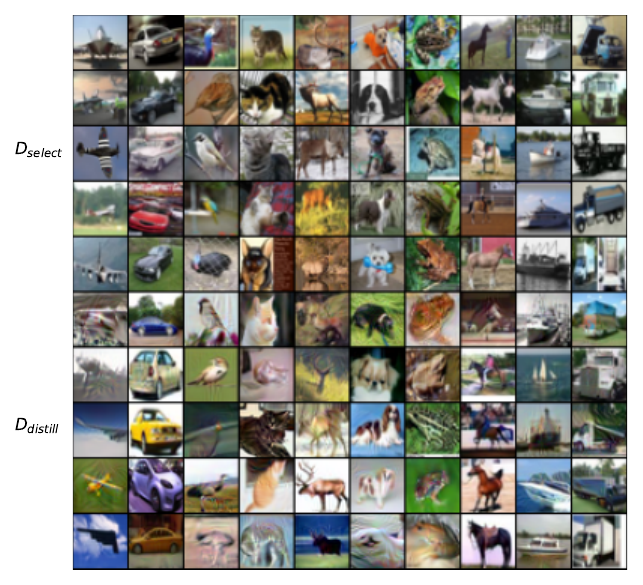}
  \end{center}
  \caption{Visualization of distilled dataset (CIFAR-10, IPC=250)}
  \label{fig:appendix_vis_cifar10_ipc250}
\end{figure*}

\begin{figure*}[htb!]
  \begin{center}
      \includegraphics[width=.95\linewidth]{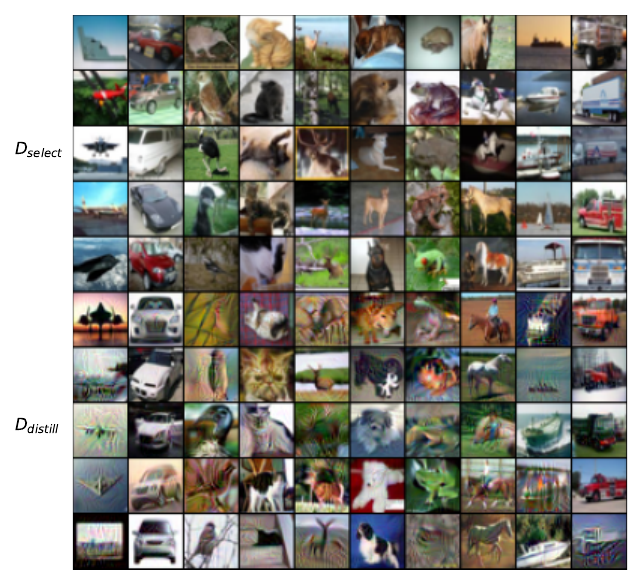}
  \end{center}
  \caption{Visualization of distilled dataset (CIFAR-10, IPC=1,500)}
  \label{fig:appendix_vis_cifar10_ipc1500}
\end{figure*}

\begin{figure*}[htb!]
  \begin{center}
      \includegraphics[width=.95\linewidth]{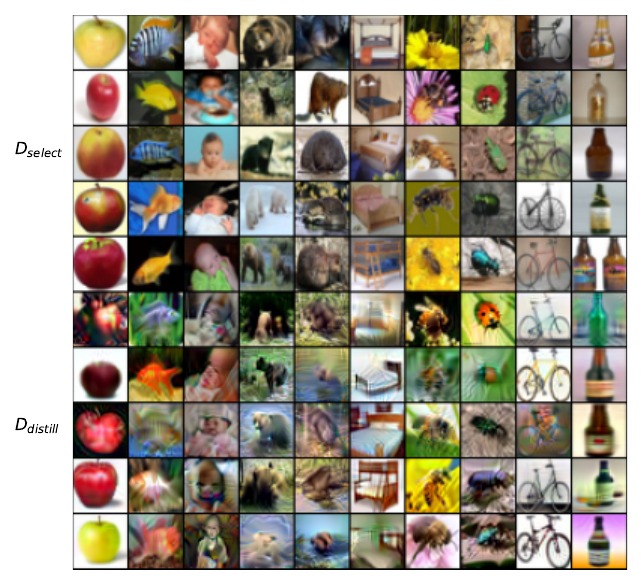}
  \end{center}
  \caption{Visualization of distilled dataset (CIFAR-100, IPC=25)}
  \label{fig:appendix_vis_cifar100_ipc25}
\end{figure*}

\begin{figure*}[htb!]
  \begin{center}
      \includegraphics[width=.95\linewidth]{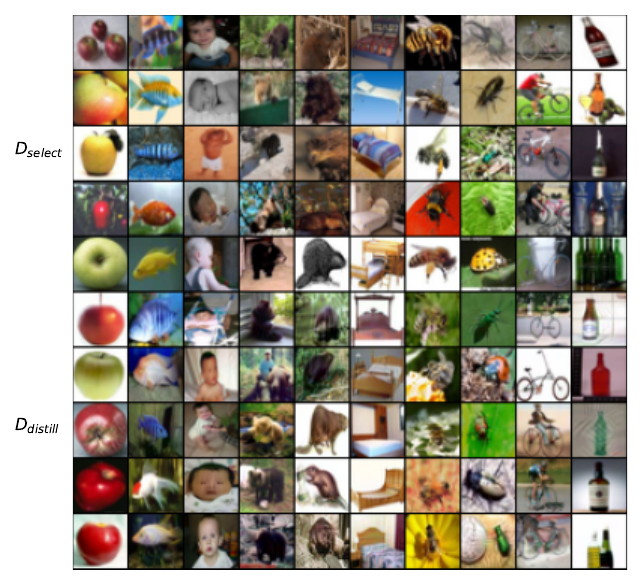}
  \end{center}
  \caption{Visualization of distilled dataset (CIFAR-100, IPC=150)}
  \label{fig:appendix_vis_cifar100_ipc150}
\end{figure*}

\begin{figure*}[htb!]
  \begin{center}
      \includegraphics[width=.95\linewidth]{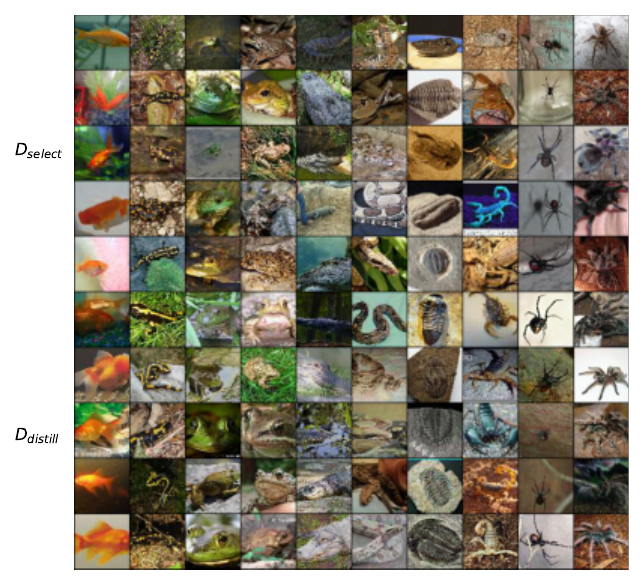}
  \end{center}
  \caption{Visualization of distilled dataset (Tiny ImageNet, IPC=50)}
  \label{fig:appendix_vis_tiny_ipc50}
\end{figure*}

\begin{figure*}[htb!]
  \begin{center}
      \includegraphics[width=.95\linewidth]{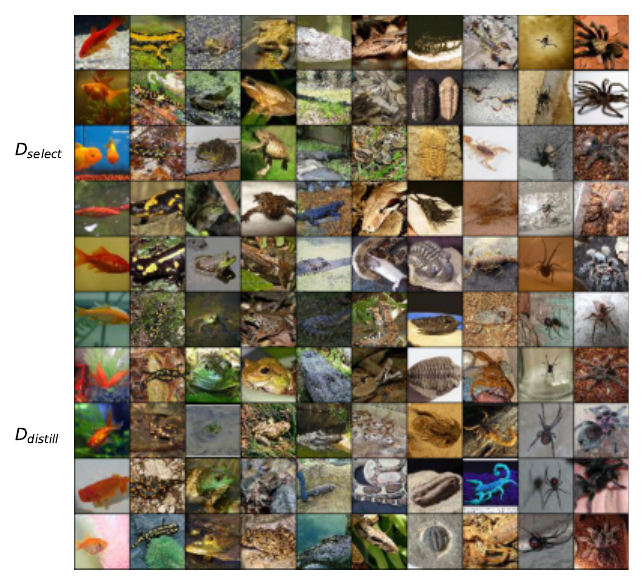}
  \end{center}
  \caption{Visualization of distilled dataset (Tiny ImageNet, IPC=100)}
  \label{fig:appendix_vis_tiny_ipc100}
\end{figure*}

We compare the synthetic images of SelMatch for CIFAR-10 with IPC=250 (ratio=5\%) and 1,500 (ratio=30\%) in Figure \ref{fig:appendix_vis_cifar10_ipc250} and \ref{fig:appendix_vis_cifar10_ipc1500}, resp., for CIFAR-100 with IPC=25 (ratio=5\%) and 150 (ratio=30\%) in Figure \ref{fig:appendix_vis_cifar100_ipc25} and \ref{fig:appendix_vis_cifar100_ipc150}, resp., and for Tiny ImageNet with IPC=50 (ratio=10\%) and 100 (ratio=20\%) in Figure \ref{fig:appendix_vis_tiny_ipc50} and \ref{fig:appendix_vis_tiny_ipc100}, respectively. In the figures, each column corresponds to 10 classes (the all 10 classes for CIFAR-10, and the first 10 classes for CIFAR-100 and Tiny ImageNet), and the top five rows are images from $\mathcal{D}_\mathrm{select}$ and the bottom five rows are images from $\mathcal{D}_\mathrm{distill}$. We can observe that the synthetic dataset generated for the larger IPC tends to include more rare and unique images than the images generated for the smaller IPC.

\end{document}